\documentclass[sigconf]{acmart}

\AtBeginDocument{%
  \providecommand\BibTeX{{%
    \normalfont B\kern-0.5em{\scshape i\kern-0.25em b}\kern-0.8em\TeX}}}

\copyrightyear{2021}
\acmYear{2021}
\setcopyright{acmcopyright}
\acmConference[KDD '21]{Proceedings of the 27th ACM
SIGKDD Conference on Knowledge Discovery and Data Mining}{August 14--18,
2021}{Virtual Event, Singapore}
\acmBooktitle{Proceedings of the 27th ACM SIGKDD Conference on Knowledge
Discovery and Data Mining (KDD '21), August 14--18, 2021, Virtual Event, Singapore}
\acmPrice{15.00}
\acmDOI{10.1145/3447548.3467187}
\acmISBN{978-1-4503-8332-5/21/08}

\usepackage{subfigure}
\usepackage{xspace}
\usepackage{multirow}
\usepackage{xcolor}
\usepackage{amsmath}
\usepackage{amsfonts}
\usepackage{bm}
\usepackage{soul}
\usepackage{algorithm}
\usepackage{balance}

\newcommand{\eat}[1]{}

\newcommand{\beijing}{{\sc Beijing}\xspace}
\newcommand{\chengdu}{{\sc Chengdu}\xspace}
\newcommand{\mugrep}{{MugRep}\xspace}

\newcommand{\eg}{\emph{e.g.},\xspace}
\newcommand{\ie}{\emph{i.e.},\xspace}

\newcommand{\etc}{\emph{etc.}\xspace}

\newcommand\figref[1]{Figure~\ref{#1}}

\newcommand\tabref[1]{Table~\ref{#1}}
\newcommand\secref[1]{Section~\ref{#1}}
\newcommand\equref[1]{Eq.~(\ref{#1})}
\newcommand{\tabincell}[2]{\begin{tabular}{@{}#1@{}}#2\end{tabular}}

\newtheorem{pro}{Problem}
\newtheorem{defi}{Definition}

\begin{document}
\fancyhead{}

\title{MugRep: A Multi-Task Hierarchical Graph Representation Learning Framework for Real Estate Appraisal}

\author{Weijia Zhang$^{1, 2\dagger}$, Hao Liu$^{2*}$, Lijun Zha$^{1}$, Hengshu Zhu$^{3}$, Ji Liu$^{2}$, Dejing Dou$^{2}$, Hui Xiong$^{4*}$}
\affiliation{
	$^1$School of Computer Science, University of Science and Technology of China,\\
	$^2$Baidu Research,
	$^3$Baidu Talent Intelligence Center, Baidu Inc.,
	$^4$Rutgers University \\
	\{wjzhang3,zlj160\}@mail.ustc.edu.com,
	\{liuhao30,zhuhengshu,liuji04,doudejing\}@baidu.com,
	hxiong@rutgers.edu
}

\thanks{
$^{*}$~Corresponding author.\\
$^{\dagger}$~The research was done when the first author was an intern in Baidu Research under the supervision of the second author.}
    

\begin{abstract}
    Real estate appraisal refers to the process of developing an unbiased opinion for real property’s market value, which plays a vital role in decision-making for various players in the marketplace~(e.g., real estate agents, appraisers, lenders, and buyers).
    However, it is a non-trivial task for accurate real estate appraisal because of three major challenges: (1)~The complicated influencing factors for property value; (2)~The asynchronously spatiotemporal dependencies among real estate transactions; (3)~The diversified correlations between residential communities. 
    To this end, we propose a \textit{\underline{Mu}lti-Task Hierarchical \underline{G}raph \underline{Rep}resentation Learning}~(\mugrep) framework for accurate real estate appraisal.
    Specifically, by acquiring and integrating multi-source urban data, we first construct a rich feature set to comprehensively profile the real estate from multiple perspectives~(\eg geographical distribution, human mobility distribution, and resident demographics distribution).
    Then, an evolving real estate transaction graph and a corresponding event graph convolution module are proposed to incorporate asynchronously spatiotemporal dependencies among real estate transactions.
    Moreover, to further incorporate valuable knowledge from the view of residential communities, we devise a hierarchical heterogeneous community graph convolution module to capture diversified correlations between residential communities. 
    Finally, an urban district partitioned multi-task learning module is introduced to generate differently distributed value opinions for real estate. 
    Extensive experiments on two real-world datasets demonstrate the effectiveness of \mugrep and its components and features.
\end{abstract}

\ccsdesc[500]{Information systems~Data mining}
\ccsdesc[500]{Information systems~Spatial-temporal systems}
\ccsdesc[500]{Applied computing~Marketing}

\keywords{Real estate appraisal; graph neural networks; multi-task learning}

\maketitle
\section{Introduction}\label{sec:intro}
Real estate appraisal provides an opinion of real property's market value, which is the probable sales price it would bring in an open and competitive real estate market.
Real estate appraisal is required and implemented by various players in the marketplace,
such as real estate agents, appraisers, property developers, investors, lenders, and buyers~\cite{pagourtzi2003real}.
An accurate appraisal of real estate is of great importance to help buyers or sellers for negotiation and closing, help mortgage owners for lending and investigation, and help governments for urban planning.

Prior studies on real estate appraisal can be categorized into two classes: (1) \textit{Empirical Appraisal Methods}~(EAMs), such as sales comparison approach~\cite{Mccluskey1997AnEO}, cost approach~\cite{guo2014integrated}, and income approach~\cite{baum2017income}, either heavily depend on the accuracy, availability, and timeliness of sale transaction data, or require strong domain knowledge to perform~\cite{pagourtzi2003real}; 
(2) \textit{Automated Valuation Methods}~(AVMs) estimate the market value of a property based on automatic machine learning techniques, such as linear regression~\cite{csipocs2008linear,ahn2012using}, support vector regression~\cite{lin2011predicting}, boosted regression trees~\cite{graczyk2010comparison, park2015using} and artificial neural networks~\cite{peterson2009neural,poursaeed2018vision,law2019take}. 
Compared with EAMs, AVMs are easy-to-use even for non-domain experts and are widely used in practice.

However, with the prevalence of mobile devices and the proliferation of ubiquitous computing techniques, existing AVMs can be significantly improved from the following three aspects.
First, existing AVMs are mainly based on fundamental real-estate information, such as the apartment attributes, residential community features, and surrounding geographical facilities~(\eg distance to mall, number of transport stations), but overlook the influence of demographic characteristics~(\eg mobility patterns and demographics of community residents), which contains important clues for real estate appraisal.
For instance, real estates in a residential community which is in close relationship~(\eg similar human mobility patterns or resident demographics) with other high-end residential communities tend to have a higher market value.
Second, existing AVMs usually regard the real estate transactions as a pile of independent samples, but ignores spatiotemporal dependencies among real estate transactions. Indeed, the value of real estate significantly depends on its spatially proximal estates~\cite{fu2014exploiting}. 
Incorporating such spatiotemporal dependencies in asynchronous real estate transactions can further improve appraisal effectiveness.
Third, the value of a real estate highly depends on the corresponding residential community.
Beyond statistic attributes, the diversified correlations among residential communities can also be exploited to achieve higher appraisal accuracy.

Along these lines, in this paper, we present a \textit{\underline{Mu}lti-Task Hierarchical \underline{G}raph \underline{Rep}resentation Learning}~(\mugrep) framework for real estate appraisal. Our major contributions can be summarized as follows:
(1) We construct a rich set of features from multi-source user data, which provides a comprehensive real estate profile from multiple social demography views,~\eg geographical population visits, human mobility, and resident demographics.
(2) By regarding each real estate transaction as an individual event, we construct an evolving transaction event graph based on both place and period proximity of transaction events, and introduce graph neural networks~(GNN) for asynchronously spatiotemporal event-level dependencies modeling to enhance appraisal effectiveness. 
To the best of our knowledge, this is the first attempt to employ GNN techniques to improve real estate appraisal.
(3) We further propose a hierarchical heterogeneous community graph convolution module to capture the diversified community-level correlations. Specifically, we devise a dynamic intra-community graph convolution block to obtain time-dependent community representation, and design a heterogeneous inter-community graph convolution block to propagate valuable knowledge from the correlated residential communities.
(4) We conduct extensive experiments on two real-world datasets, the results demonstrate the effectiveness of our framework and its components and features.

\begin{table}[tb]
	\small
	\centering
	\caption{Statistics of datasets.}
	\vspace{-2.5mm}
	\setlength{\tabcolsep}{1mm}{
		\begin{tabular}{l|l|rr}
			\toprule[0.8pt]
			\multicolumn{2}{c|}{\textbf{Description}} & \beijing & \chengdu\\
			\midrule[0.5pt]
			\multirow{2}{*}{\tabincell{l}{Real Estate \\Marketing Data}} & \# of transactions & 185,151 & 134,781 \\
			~ & \# of communities & 6,267 & 3,995 \\
			\midrule[0.5pt]
			\multirow{2}{*}{\tabincell{l}{Geographical \\Data}} & \# of POIs & 1,315,353 & 1,116,989\\
			~ & \tabincell{l}{\# of transport stations} & 44,513 & 30,564\\
			\midrule[0.5pt]
			Check-in Data & \# of check-ins & 4,078,723,999 & 5,130,847,589 \\
			\midrule[0.5pt]
			User Trip Data & \# of trip queries & 73,307,426 & 66,960,348 \\
			\midrule[0.5pt]
			User Profile Data & \# of users & 2,042,718 & 1,426,860 \\
			\bottomrule[0.5pt]
	\end{tabular}}
	\vspace{-3mm}
	\label{table:dataset}
\end{table}

\begin{figure*}[tb]
	\centering
    \vspace{-6mm}
    \hspace{-4.5mm}
	\subfigure[{Unit price distribution.}]{\label{fig:unit_price}
		\includegraphics[width=0.68\columnwidth]{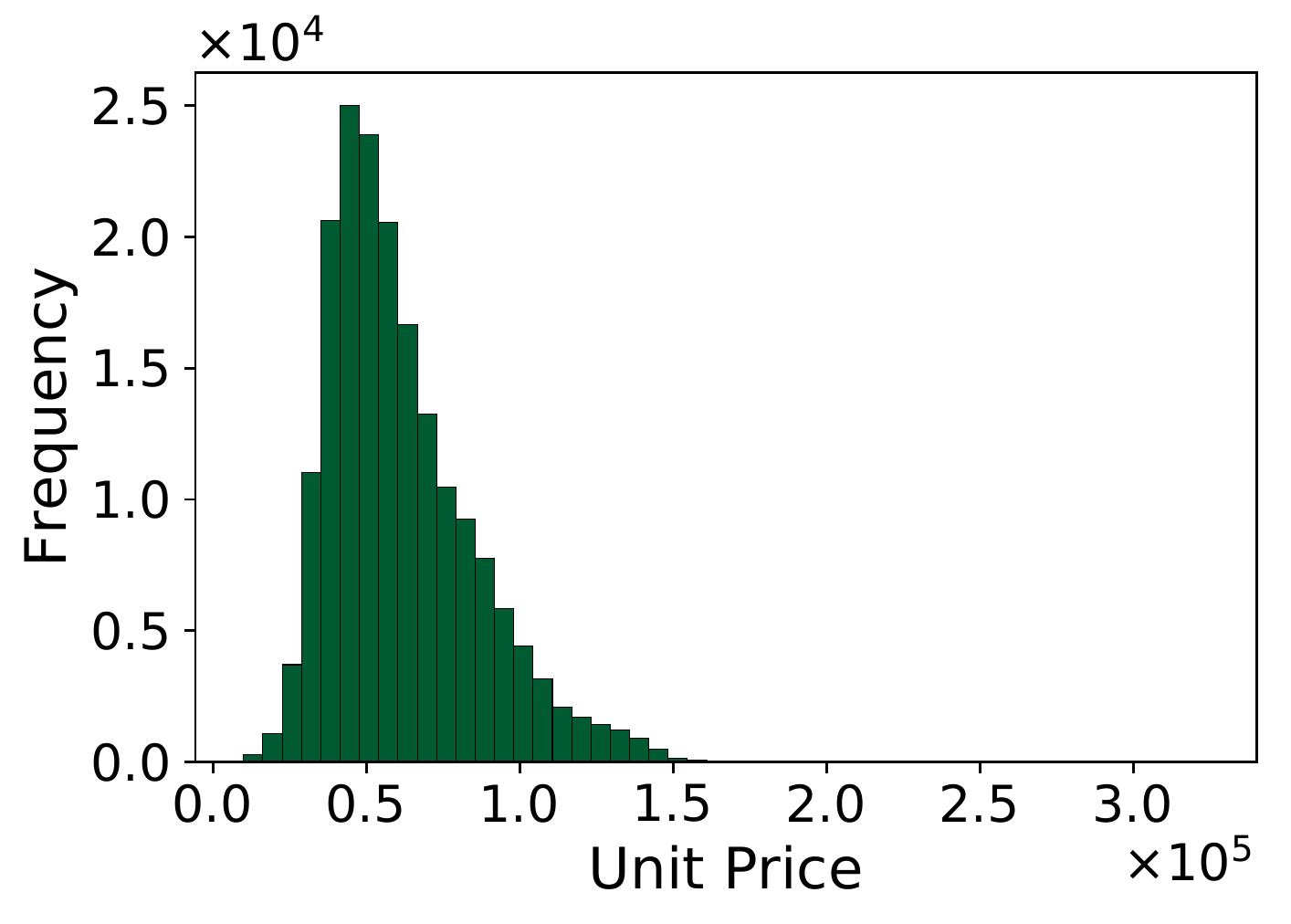}}
	\subfigure[{Temporal distribution of unit price.}]{\label{fig:price_month}
		\includegraphics[width=0.68\columnwidth]{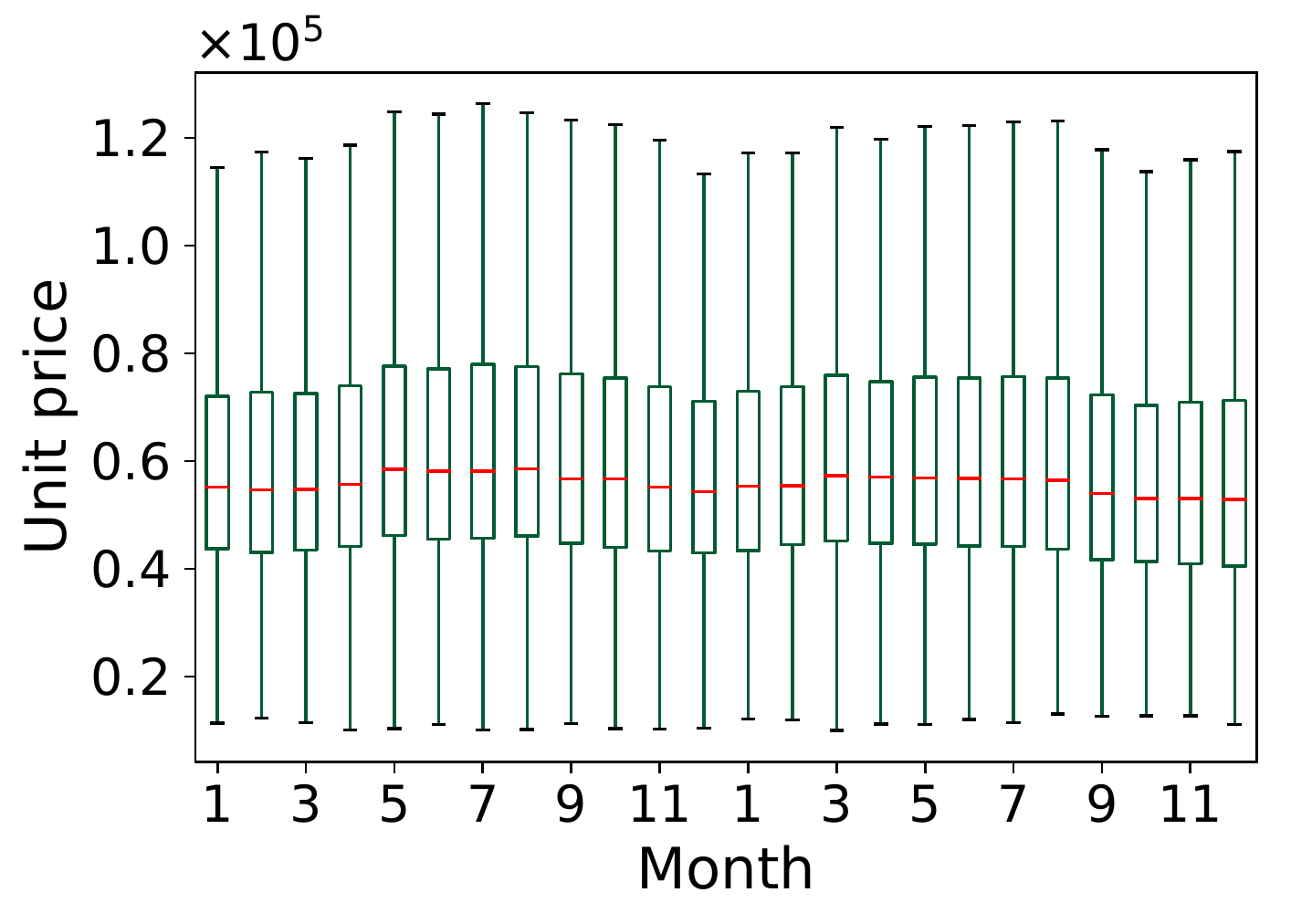}}
	\subfigure[{Temporal distribution of transaction volume.}]{\label{fig:ntrans_month}
		\includegraphics[width=0.68\columnwidth]{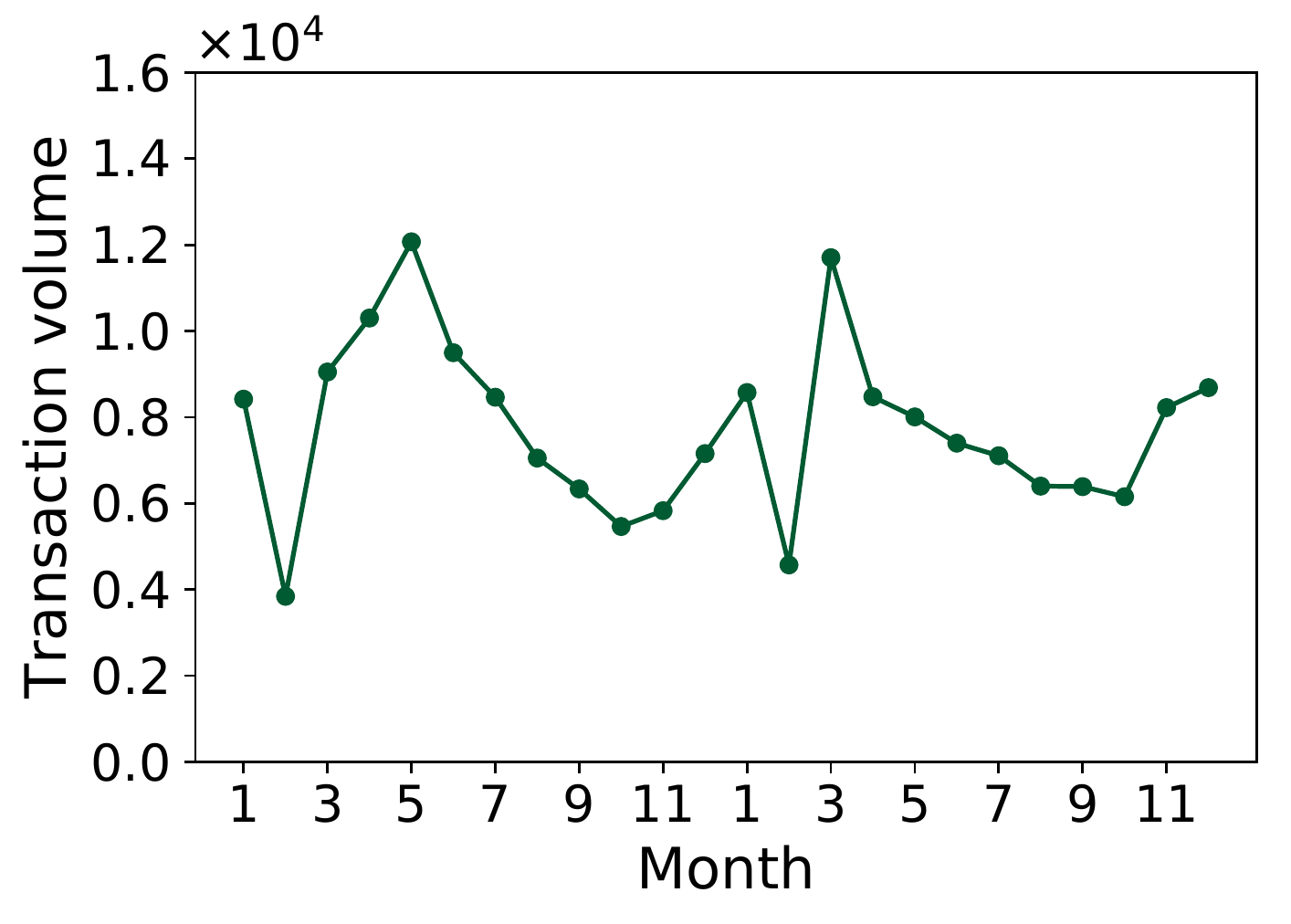}}
	\subfigure[{Spatial distribution of unit price.}]{\label{fig:price_dist}
		\includegraphics[width=0.58\columnwidth]{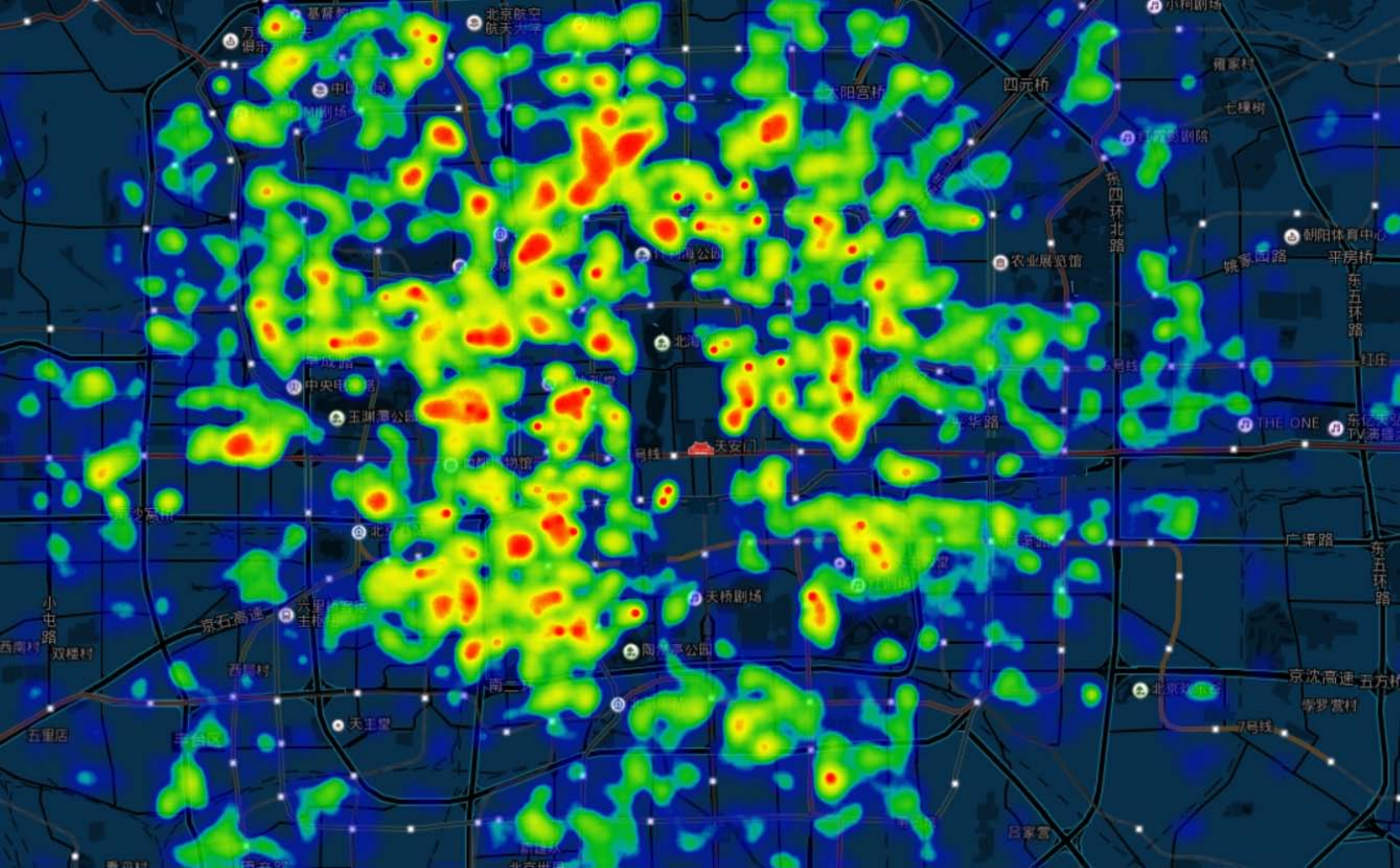}}
		\hspace{8mm}
	\subfigure[{Spatial distribution of transaction volume.}]{\label{fig:ntrans_dist}
		\includegraphics[width=0.58\columnwidth]{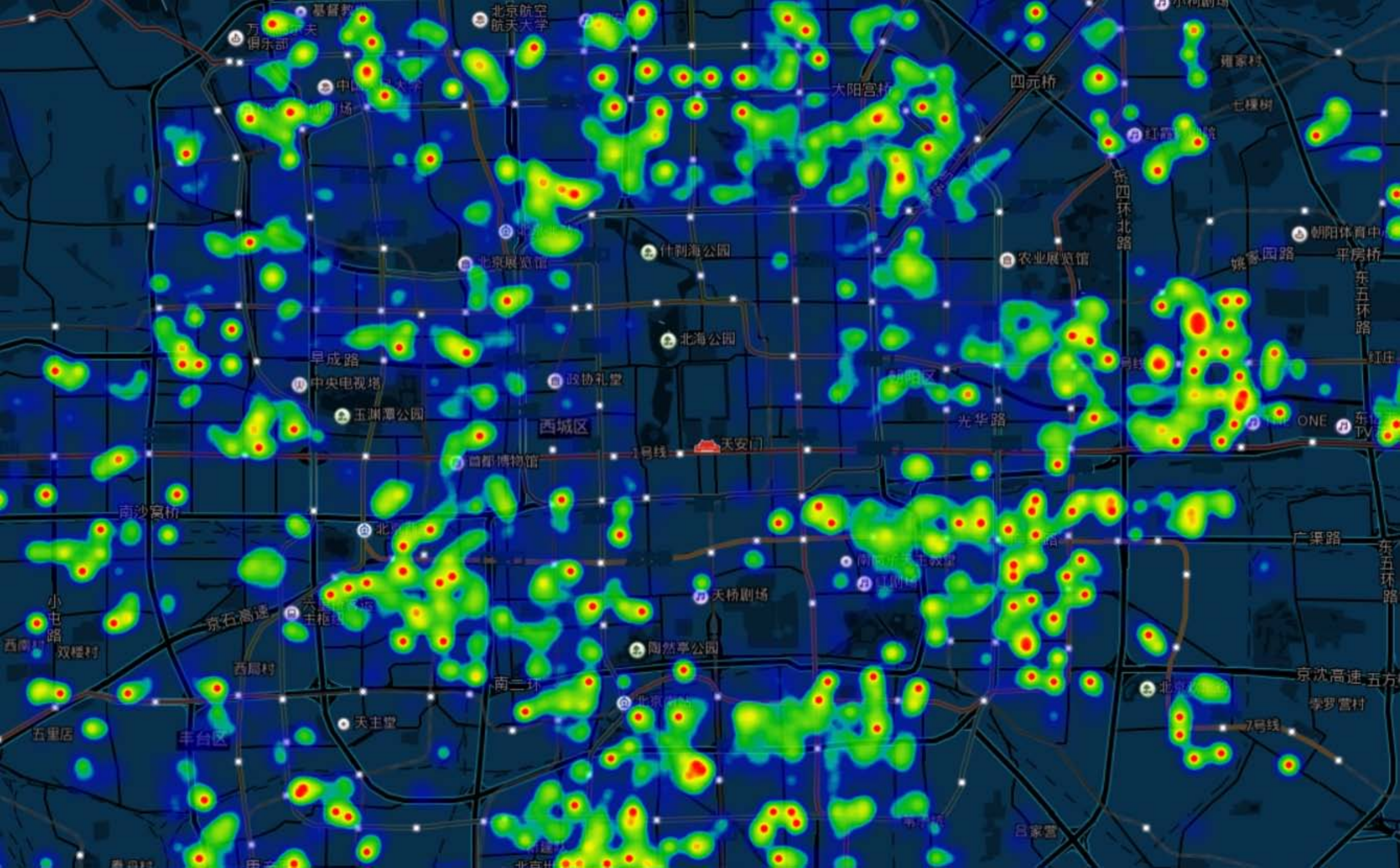}}
	\hspace{8mm}
	\subfigure[{Spatial distribution of residential community.}]{\label{fig:community_dist}\includegraphics[width=0.58\columnwidth]{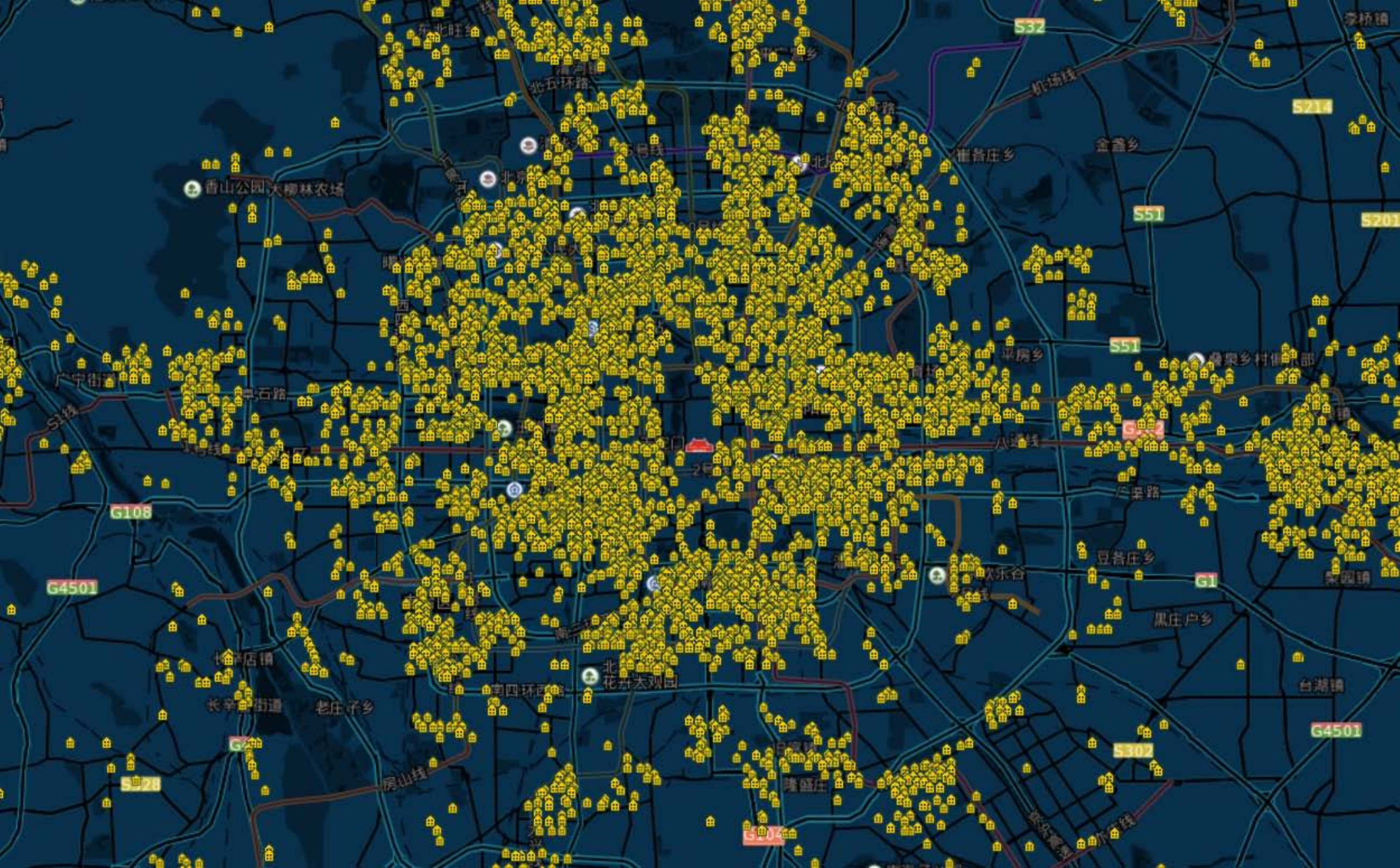}}
	\hspace{-4.5mm}
	\vspace{-2mm}
	\caption{Analysis and visualization on \beijing.} 
	\vspace{-3mm}
	\label{fig:basic_data}
\end{figure*}

\section{Data description and analysis}
In this section, we present the datasets to be used in our framework, with a preliminary data analysis. We use two datasets, \ie~\beijing and \chengdu, which represent two major metropolises in China. 
Except basic \textit{Real Estate Marketing Data}, we exploit four additional multi-source urban datasets, \ie \textit{Geographical Data}, \textit{Check-in Data}, \textit{User Trip Data}, and \textit{User Profile Data}, to improve the accuracy of real estate appraisal. 
\tabref{table:dataset} summarizes the statistics of the datasets.

\subsection{Real Estate Marketing Data}
In this paper, the real estate marketing datasets of \beijing and \chengdu are collected from a major commercial real estate agency\footnote[1]{https://bj.lianjia.com}. 
The scope of \beijing ranges from January 2018 to December 2019, and that of \chengdu ranges from January 2018 to December 2020. Each dataset consists of two kinds of data, \ie \textit{real estate transaction data} and \textit{residential community data}.  

We take \beijing as an example. 
\beijing totally contains 185,151 real estate transaction records and 6,267 residential communities. Each transaction record contains the attributes of the transaction estate, \eg transaction date, transaction price, location, room number, area, transaction ownership, whether it is free of sales tax, the residential community real estate belongs to. \figref{fig:unit_price} shows the distribution of real estate unit price in \beijing. \figref{fig:price_month} and \figref{fig:ntrans_month} show the fluctuation of unit price and transaction volume over time.  \figref{fig:price_dist} and \figref{fig:ntrans_dist} show the spatial distributions of unit price and transaction volume. Overall, the variation of real estate unit price in spatial domain is greater than that in temporal domain, whereas the transaction volume is distributed more evenly across the city with a notable periodical pattern.
For each residential community, our datasets contain its developer, completion year, number of estates, property fee, \etc
\figref{fig:community_dist} shows the spatial distribution of residential communities in \beijing, which is positively correlated with spatial distribution of transaction volume.

\subsection{Geographical Data}
Then we construct large-scale geographical datasets, including \textit{point of interest (POI) data}~\cite{li2020competitive} and \textit{transport station data}~\cite{liu2020incorporating}. 
There are 1,315,353 POIs and 44,513 transport stations in \beijing, 1,116,989 POIs and 30,564 transport stations in \chengdu.

\subsection{Check-in Data}
Each check-in record corresponds to a GPS request from a mobile user, which is collected through Baidu location SDK~\cite{zhang2020semi,zhu2020rapid}.
There are 4,078,723,999 and 5,130,847,589 users' check-ins in \beijing and \chengdu respectively, to support the real estate appraisal task.

\subsection{User Trip Data}
The \textit{User Trip Data} are collected from Baidu Maps, which records the mobility pattern of a city. The user trip data includes the origin~(geographical location) and destination of a trip, user's travel mode~(\eg drive, taxi, bus, cycle, walk), and the type~(\eg enterprise, shopping places) of trip destination.
Overall, there are 73,307,426 and 66,960,348 trip records in \beijing and \chengdu, respectively.

\subsection{User Profile Data}
The \textit{User Profile Data} contain user profile attributes~(\eg gender, age, income level, education level), which is collected from multiple Baidu applications including Baidu Search, Baidu App and Baidu Maps. 
There are 2,042,718 and 1,426,860 distinct user records in \beijing and \chengdu, respectively. Each record contains a user's demographic attributes including hometown, gender, age, and social attributes such as the industry, income level, educational level, consumption level, and whether the user is a car owner. All user profile records are anonymous and cannot be associated with sensitive personal information such as names and phone numbers.

\section{Preliminaries}\label{sec:preliminaries}
We first introduce some important definitions and formally define the real estate appraisal problem. 
\begin{defi}
	\textbf{Subject Property}.
	The subject property is the real estate that will be appraised. 
\end{defi}
\begin{defi}
	\textbf{Real Estate Transaction Event}.
	Consider a set of real estate transactions $S$, a real estate transaction event $s_t =\langle l_t, T_t, x_t, y_t \rangle \in S$ is defined as the $t$-th chronological real estate transaction in the dataset. 
	Specifically, $l_t$ is the location of $s_t$, $T_t$ is transaction date, \eat{$x_t=[f_{h,t},f_{c,t}^i,f_{g,t}^i,f_{v,t}^i,f_{m,t}^i,f_{p,t}^i]$}$x_t$ is the feature input associated with the real estate in $s_t$, and $y_t$ is the transaction unit price of $s_t$. 
\end{defi}
\begin{pro}
    \textbf{Real Estate Appraisal}.
	Given the subject property $s_{t+1}$, our task is to estimate the transaction unit price of $s_{t+1}$. 
\end{pro}

\section{Framework}
Then we introduce our framework in detail, including the processes of feature construction, the event-level and community-level representation learning, and multi-task learning based valuation.

\subsection{Overview}
\begin{figure}[tb]
	\centering
    \vspace{1.5mm}
	\includegraphics[width=1\columnwidth]{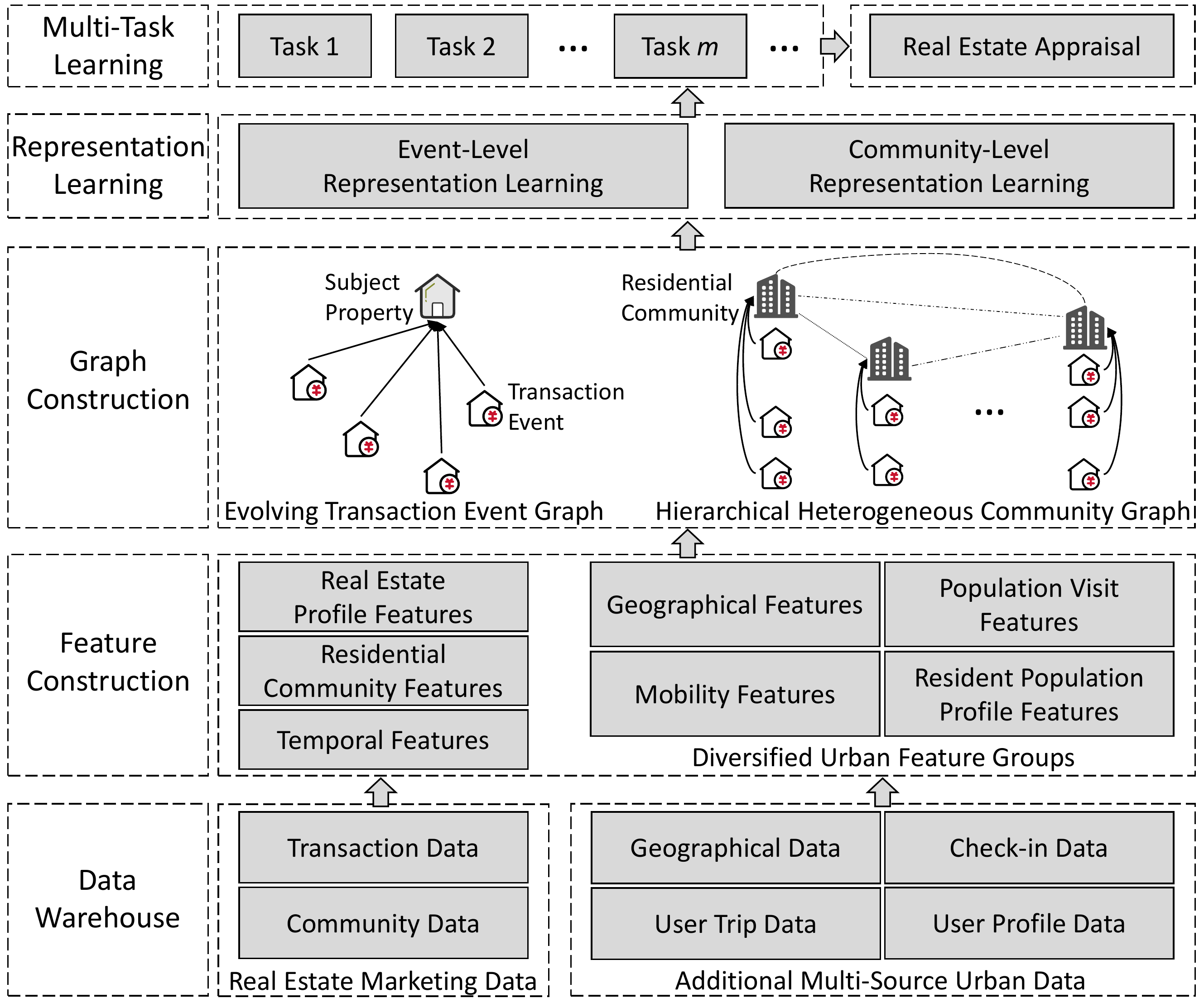}
	\caption{The framework overview of \mugrep.}
	\label{fig:framework}
	\vspace{-3mm}
\end{figure}

\figref{fig:framework} shows the framework overview of \mugrep, which consists of five major components: Data Warehouse, Feature Construction, Graph Construction, Representation Learning, and Multi-Task Learning. The Data Warehouse stores \textit{Real Estate Marketing Data} and four additional multi-source urban data. The Feature Construction module generates seven groups of features based on the corresponding datasets in Data Warehouse. 
Besides, the Graph Construction module builds evolving transaction event graph based on the place and period proximity of transaction events, and constructs hierarchical heterogeneous community graph based on intra-community real estate transactions and inter-community similarities. In particular, the community similarities are quantified by additional diversified urban feature groups. 
After that, the Representation Learning module is introduced to derive the event-level and community-level representation of subject property, of which the integration is processed in a Multi-Task Learning module, where each task corresponds to an urban district. The corresponding task block produces the final result of Real Estate Appraisal. 

\subsection{Feature Construction}\label{sec:feat_construct}
In this subsection, we present the process of constructing various features based on the aforementioned multi-source urban datasets.
The details of used features are listed in \tabref{table:features} of Appendix \ref{sec:feature_detial}.

\subsubsection{\textbf{Real Estate Profile Features.}}
The real estate's profiles are the most fundamental factors for the real estate appraisal.
We extract real estate profile features from \textit{real estate transaction data}, including estate's number of rooms, area, decoration, orientation, structure, free of tax, transaction ownership, \etc
The features of estate also include some profiles of the building where it is located, such as floor number, building type, elevator household ratio.

\subsubsection{\textbf{Residential Community Profile Features.}}
The residential community where the estate is located is another crucial factor that determines the value of estate. The residential community profile features include its developer, completion year, number of buildings and estates, property fee and district. All the above features are extracted from \textit{residential community data}. In addition, we identify each community by a unique identifier.

\subsubsection{\textbf{Temporal Features.}}
The temporal features include the valuation date of estate and the price distribution of historical transactions of the same residential community.
Historical estate transaction prices in the same community are important reference factors for subject property appraisal, because these estates usually have many similar attributes~(\eg community attributes, geographical attributes), therefore, have similar price distribution. We quantify the distribution of historical prices by some statistic features~(\eg mean, variance, maximum, minimum) for the unit prices of estate transactions that have been closed in previous quarter of the valuation date in the same residential community.

\subsubsection{\textbf{Geographical Features.}}
The geographical features are of much importance when appraising a real estate as well. 
The estate in an area that has complete facilities, which provide more convenience and enjoyment for living, usually has a higher price.
Except counting the number of all POIs and transport stations close to the estate to reflect the completeness of surrounding facilities, we further consider several kinds of important geographical factors, including transportation, education, medical treatment, shopping, living, entertainment, and unpleasantness. We count the number of facilities and places that correspond to above factors nearby the estate, and calculate the distance from the estate to the nearest ones. 
The spatial distribution of facilities number feature is shown in \figref{fig:poi_dist}, where we can observe a positive correlation between this feature and real estate unit price.
More details of geographical features can be found in \tabref{table:features}.

\subsubsection{\textbf{Population Visit Features.}}
The geographical visitation volume can reflect the popularity and prosperity of an area, which has a significant impact on real estate prices.
We first aggregate the check-in records of each user by every 10 minutes as one visit of a location. Then we construct population visit features by accumulating the visiting frequency of population nearby the estate in work hours~(10:00-18:00), break hours~(18:00-23:00), and all day on workdays and weekends, respectively.
By comparing \figref{fig:price_dist} and \figref{fig:visit_dist}, we observe the areas with high visiting frequency of population are usually of high unit prices of real estates.

\subsubsection{\textbf{Mobility Features.}}
Human mobility is also an important factor to estimate the real estate prices of a residential community. For example, if residents of a community frequently visit high-consumption or high-income places, 
then we can infer higher real estate prices for the residential community.
Thus, we construct abundant mobility features based on \textit{User Trip Data} to reflect mobility patterns of community residents, including the human volume of inflow and outflow of community, the distribution of travel modes~(e.g., driver, taxi, bus, cycle, walk) and the distribution of travel destination types~(e.g., enterprise, administration, shopping places, entertainment venues) of community residents on workdays and weekends, respectively.

\subsubsection{\textbf{Resident Population Profile Features.}}
\begin{figure*}[tb]
	\centering
    \vspace{-4mm}
	\subfigure[{Spatial distribution of facilities number.}]{\label{fig:poi_dist}
		\includegraphics[width=0.58\columnwidth]{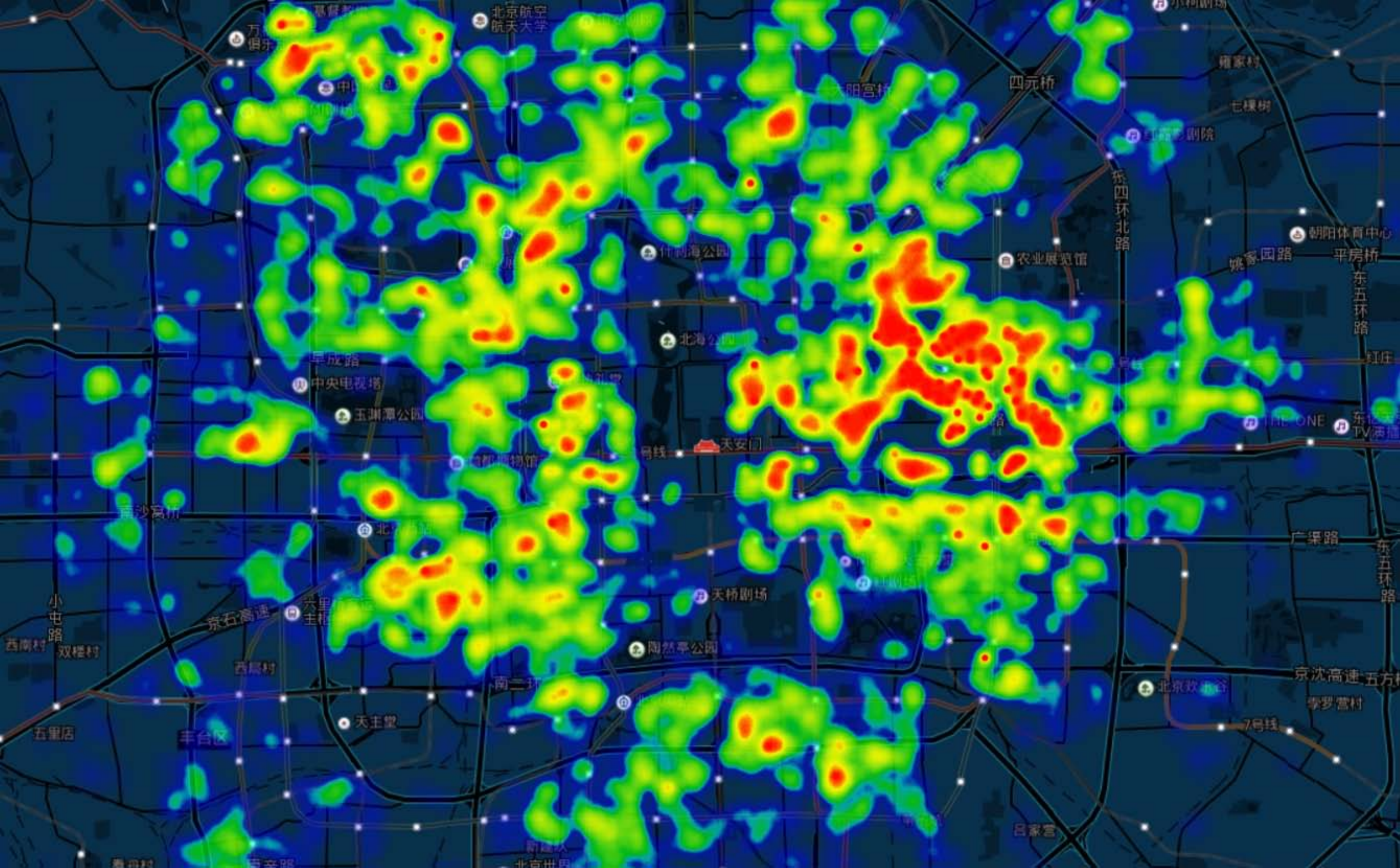}}
	\hspace{8mm}
	\subfigure[{Spatial distribution of population visits.}]{\label{fig:visit_dist}
		\includegraphics[width=0.58\columnwidth]{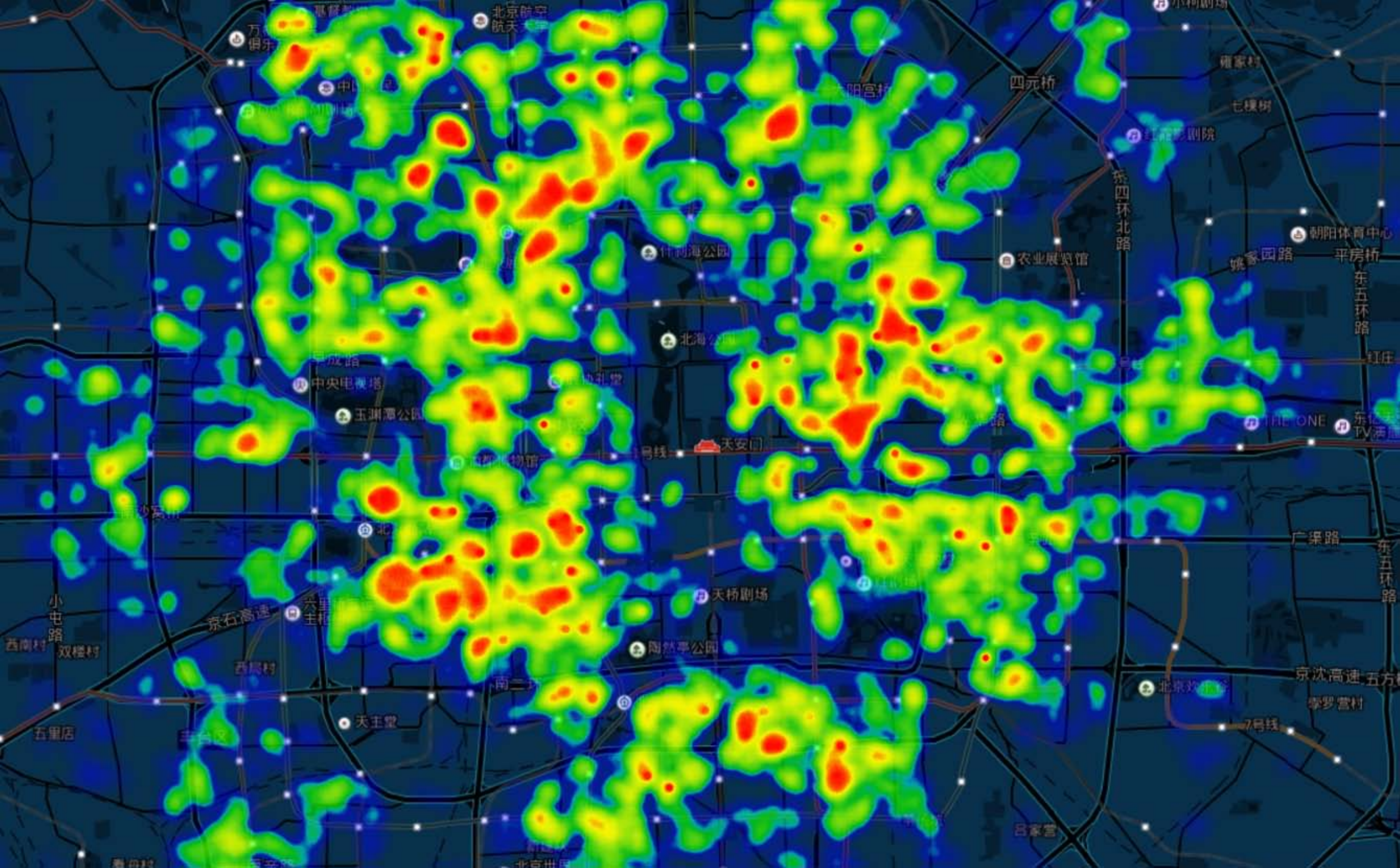}}
	\hspace{8mm}
	\subfigure[{Spatial distribution of residents' incomes.}]{\label{fig:income_dist}
		\includegraphics[width=0.58\columnwidth]{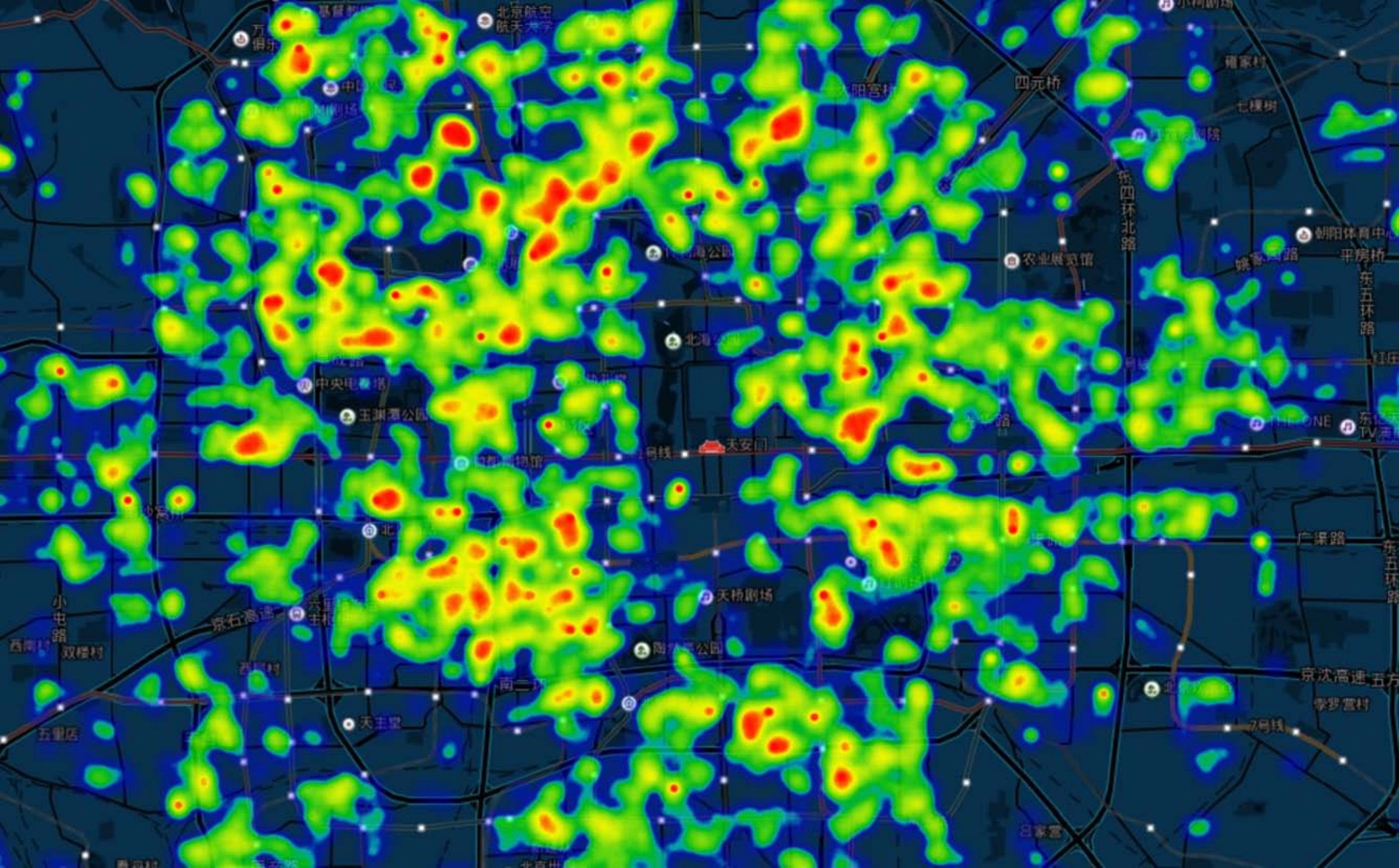}}
	\vspace{-2mm}
	\caption{Spatial distributions of features on \beijing.} 
	\vspace{-3mm}
	\label{fig:basic_feature}
\end{figure*}

The resident population profile also has strong correlation with real estate prices of the residential community. For example, the residential communities where high-income and high-consumption residents live indicates the brownstone districts, and are supposed to have high real estate prices. 
To this end, we construct valuable resident population profile features of the community based on \textit{User Profile Data} to comprehensively profile residents' demographic attributes and social attributes. These features include the resident population number, the distributions of residents' hometown, life stage, industry, income level, education level, consumption level, \etc
The details of these features are listed in \tabref{table:features}.
We depict the spatial distribution of community residents' income in \figref{fig:income_dist}, which further confirms that the communities with high-income residents usually correspond to high real estate prices.

\subsection{Event-level Representation Learning}\label{sec:event_learning}
The price of real estate transactions in proximal places and periods have strong dependencies. For example, for a place with a planned subway station, the real estate prices of surrounding areas usually increase synchronously. The transaction price dependencies can also be validated by \figref{fig:price_dist},~\ie the nearby real estate prices tend to be similar. 
However, these transactions are dispersedly distributed in spatial and temporal domains, which induces spatiotemporal asynchronism. 
Such asynchronously spatiotemporal dependencies among real estate transactions also distinguish our task from existing works on predicting regional future house prices~\cite{tan2017time,ge2019integrated}, where the input are more regular time series data. It also prevents us to adopt existing spatiotemporal prediction approaches~\cite{li2018dcrnn_traffic} for our task.

To tackle above problem, we first formulate each real estate transaction as a transaction event, which is defined in \secref{sec:preliminaries}.
Then, we model the continuously emerging transaction events as an evolving graph $G_{e}=(V_{e}, E_{e}, A_{e})$, where $V_{e}=S$ is a set of real estate transaction events, $E_{e}$ is a set of edges indicating connectivity among transaction events, and $A_e$ denotes the proximity matrix of $G_e$. Specifically, we define connectivity constraint $e_{(t+1)t'} \in E_e$ as
\begin{equation}
e_{(t+1)t'}=\left\{
\begin{aligned}
&1,\quad dist(s_{t+1},s_{t'}) \leq \epsilon_d,\  0<(T_{t+1}-T_{t'}) \leq \epsilon_{\tau}  \\
&0,\quad otherwise
\end{aligned},
\right.
\end{equation}
where $dist(\cdot)$ is the geographical distance between the locations of events, $\epsilon_d$ and $\epsilon_{\tau}$ are physical distance threshold and time interval threshold, respectively. To balance the connected events from different communities, we restrict a node at most to connect last $N_e$ events from each community.
With real estate transaction events occurring over time, the graph $G_e$ evolves accordingly.

Inspired by recent success of GNN~\cite{kipf2017semi,velivckovic2017graph} on processing non-Euclidean graph structures and its inductive ability to effectively generate representation for previous unseen node~\cite{hamilton2017inductive}, we adopt graph attention networks~\cite{velivckovic2017graph}, an effective variant of GNN, to capture event-level dependencies among real estate transactions on the evolving graph.
	
Specifically, to quantify the sophisticated influence of historical transaction events $s_{t'}$ to subject property $s_{t+1}$, we
introduce the attention mechanism~\cite{zhang2021intelligent} to automatically learn the coefficient between transaction events,
\begin{equation}
\beta_{(t+1)t'} = \mathbf{v}_e^{\top}\text{tanh}\left(\mathbf{W}_e[x_{t+1} \oplus x_{t'} \oplus y_{t'} ]\right),
\end{equation}
where $\mathbf{v}_e$ and $\mathbf{W}_e$ are learnable parameters, $\oplus$ denotes the concatenation operation. The proximity weight between event $s_{t+1}$ and $s_{t'}$ is further defined as
\begin{equation}
\label{equ:alpha}
\alpha_{(t+1)t'} = \frac{exp(\beta_{(t+1)t'})}{\sum_{k \in \mathcal{N}_{t+1}}exp(\beta_{(t+1)k})},
\end{equation}
where $\mathcal{N}_{t+1}$ is a set of adjacent events of $s_{t+1}$ in $G_{e}$.

Once $\alpha_{(t+1)t'} \in A_e$ is obtained, we derive the representation that integrates nearby previous transaction events by aggregating and transforming the adjacent events of subject property, defined as
\begin{equation}
h_{e,t+1}^{l} = \text{ReLU} \left(\mathbf{W}_{he}^l \left(\sum_{{t'}\in \mathcal{N}_{t+1}} \alpha_{(t+1)t'} {h}_{e,t'}^{l-1} + I(l>1){h}_{e,t+1}^{l-1}\right)\right),
\end{equation}
where $l$ indicates the $l$-th graph convolution layer in evolving graph, $\mathbf{W}_{he}^l$ are learnable parameters for $l$-th graph convolution, $h_{e,t+1}^{0}=x_{t+1}$, and
$I(l>1)$ is an indicator function that equals one if $l > 1$ and zero otherwise.
By stacking $L_e$ convolution layers, we can aggregate $L_e$-hop adjacent events to mine spatio-temporally extensive knowledge from historical real estate transaction data for more accurate subject property valuation. 

\subsection{Community-level Representation Learning}
As aforementioned, the real estate value is also highly correlated with the residential community it belongs to~\cite{fu2015real}.  
Therefore, an expressive representation of community can be useful in real estate appraisal. 
In this work, we devise a \textit{hierarchical heterogeneous community graph convolution} module, including the \textit{dynamic intra-community graph convolution} block and the \textit{heterogeneous inter-community graph convolution} block.
\subsubsection{\textbf{Dynamic Intra-Community Graph Convolution}}
The representation of a residential community $c^i$ should be updated dynamically once there is a new transaction event, defined as $s_t^i$, happening within it. The transaction events happened in each community $c^i$ can make up an individual impact graph: $G^i=(V^i, E^i, A^i)$, where $V^i$ consists of the community $c^i$ and transaction events happening in $c^i$, $E^i$ is a set of edges constraining what transaction events have impacts on the representation of $c^i$, and $A^i$ denotes the impact matrix of $G^i$, indicating the impacts of previous transaction events to the community representation. The connectivity $e_{t'}^i \in E^i$ between community $c^i$ and transaction event $s_{t'}^i$ is defined as
\begin{equation}
e_{t'}^i=\left\{
\begin{aligned}
&1,\quad 0\leq(T_t-T_{t'})\leq max(\epsilon_{\tau},D^i_{t,N_c})  \\
&0,\quad otherwise
\end{aligned},
\right.
\end{equation}
where $D^i_{t,N_c}$ denotes the number of days from the last $N_c$-th transaction event to the latest transaction event $s_t^i$ in $c^i$. Note the transaction event nodes set and the edges set connecting between community and transaction events change dynamically as the new transaction events occur. 

Then the attention mechanism is further used to quantify the impact of each previous transaction within $c^i$,
\begin{equation}
\beta_{t'}^i = \mathbf{v}_u^{\top}\text{tanh}\left(\mathbf{W}_u[x_{t'}^i \oplus y_{t'}^i ]\right),
\end{equation}
where $\mathbf{v}_u$ and $\mathbf{W}_u$ are learnable parameters. 
Similar to \equref{equ:alpha}, we can finally derive the impact weight $\alpha_{t'}^i$.

Once $\alpha_{t'}^i \in A^i$ is obtained, the representation of each community $c^i$ is updated by performing the graph convolution operation
\begin{equation}
h_{u}^i = \text{ReLU} \left(\mathbf{W}_{hu} \left(\sum_{{t'}\in \mathcal{N}_{i}} \alpha_{t'}^i {x}_{t'}^i \right)\right),
\end{equation}
where $\mathbf{W}_{hu}$ are learnable parameters.

\subsubsection{\textbf{Heterogeneous Inter-Community Graph Convolution}}
The diversified correlations between residential communities are induced by their various similarities.
For example, the residential communities located in similar functional areas with similar surrounding geographical facilities distribution usually tend to be positively correlated in real estate value.
Thus, we define four similarity metrics between residential communities based on four types of features,~\ie geographical features, population visit features, mobility features, and resident population profile features. 

Then, we construct the heterogeneous inter-community graph to model the diversified correlations between residential communities, which is defined as $G_{c}=(V_{c}, E_{c}, A_{c})$, where $V_{c}$ is a set of residential communities. We construct four types of edges $E_{c}=\{\mathcal{E}_{g},\mathcal{E}_{v},\mathcal{E}_{m},\mathcal{E}_{p}\}$ based on four kinds of similarities between residential communities. 
Next, we will take geographical edges set $\mathcal{E}_{g}$ as a representative for detailed explanation.

We define the geographical features of community $c^i$ as $f^i_g$.
Then, we can compute the euclidean distance between $f^i_g$ and $f^j_g$
\begin{equation}
dist_g(c^i,c^j) = \lVert f^i_g-f^j_g \rVert.
\end{equation}
Smaller euclidean distance of features indicates higher similarity between communities. Thus, the geographical edge $e_{ij}^g \in \mathcal{E}_{g}$ is defined as 
\begin{equation}
e_{ij}^g=\left\{
\begin{aligned} &1,\quad dist_g(c^i,c^j) \leq \epsilon_g\\
&0,\quad otherwise
\end{aligned},
\right.
\end{equation}
where $\epsilon_g$ is the distance threshold. 
Other types of edge sets can be derived in the same way but with respective thresholds.

We define $s_{t+1}^i$ as the subject property belonging to community $c^i$. 
With the latest representation of each community and the heterogeneous edges set $E_c$, the coefficient between communities $c^i$ and $c^j$ is computed by
\begin{equation}
\beta_{ij} = \mathbf{v}_c^{\top}\text{tanh}\left(\mathbf{W}_c[x_{t+1}^i \oplus h_u^j \oplus p_{ij}]\right),
\end{equation}
where $\mathbf{v}_c$ and $\mathbf{W}_c$ are learnable parameters, $p_{ij}$ is a one-hot vector to denote the type of edge. 
Once coefficients are derived, the proximity weight $\alpha_{ij} \in A_c$ can be obtained similar to \equref{equ:alpha}.
Then, we derive the overall representation of residential communities by the graph convolution operation
\begin{equation}
h_{c}^{i,l} = \text{ReLU} \left(\mathbf{W}_{hc}^l \left(\sum_{{j}\in \mathcal{N}_i} \alpha_{ij} {h}_{c}^{j,l-1} + I(l>1){h}_{c}^{i,l-1}\right)\right),
\end{equation}
where $l$ indicates the $l$-th graph convolution layer in inter-community graph, $\mathbf{W}_{hc}^l$ are learnable parameters for $l$-th graph convolution, ${h}_{c}^{i,0}=h_u^i$.
By stacking $L_c$ convolution layers, $h_{c}^{i,L_c}$ can integrate $L_c$-hop diversified correlations between communities.

Finally, we obtain the overall representation of subject property $s_{t+1}^i$ through concatenation and multi-layer perceptron operations
\begin{equation}
h_{t+1}^o = MLP([x_{t+1}^i \oplus h_{e,t+1}^{L_e} \oplus h_{c}^{i,L_c}]).
\end{equation}

\subsection{Multi-Task Learning Based Valuation}
\begin{figure}[tb]
	\centering
	\vspace{-1mm}
	\subfigure[{Xicheng district.}]{
		\includegraphics[width=0.48\columnwidth]{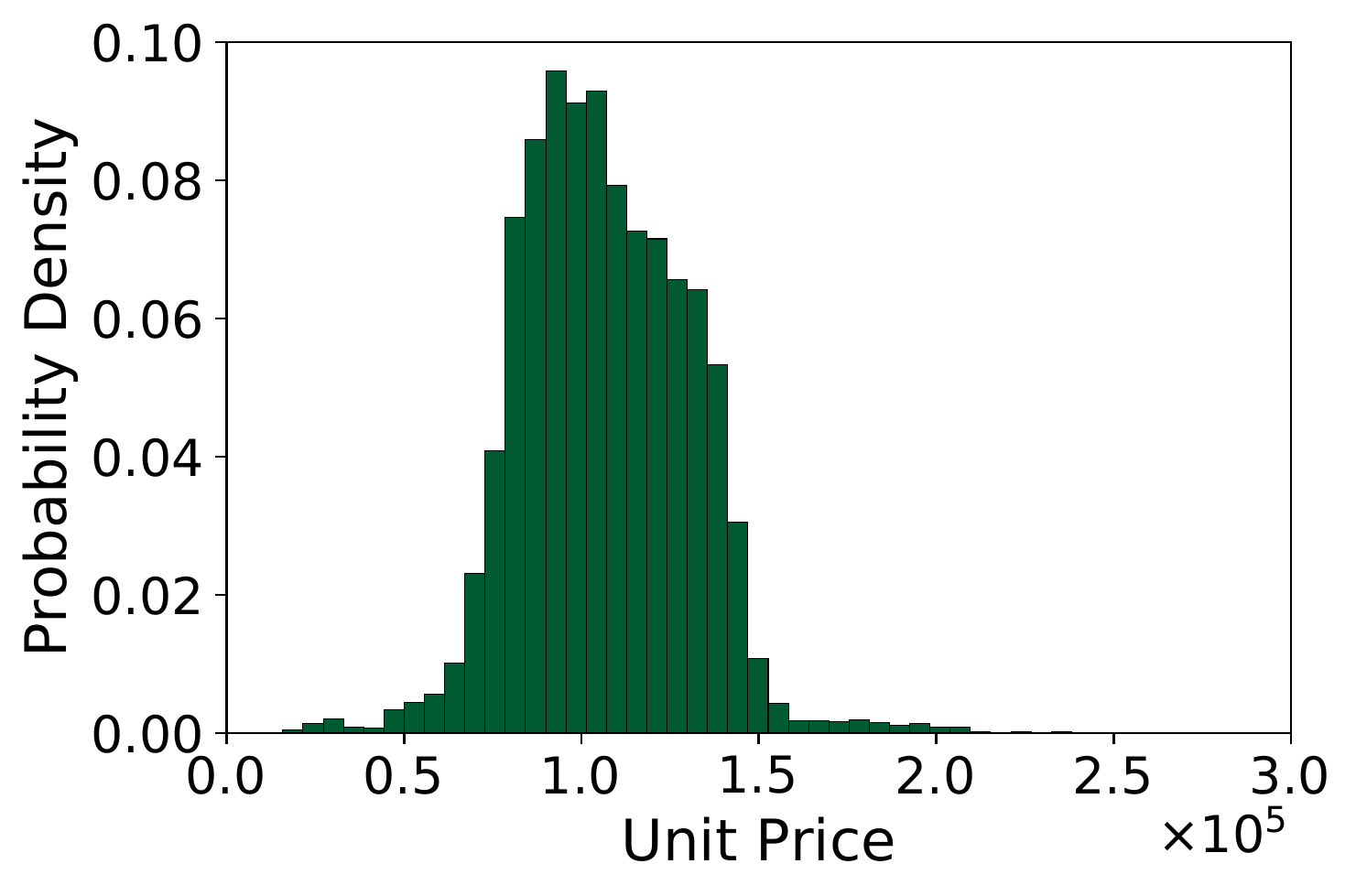}}
	\subfigure[{Changping district.}]{
		\includegraphics[width=0.48\columnwidth]{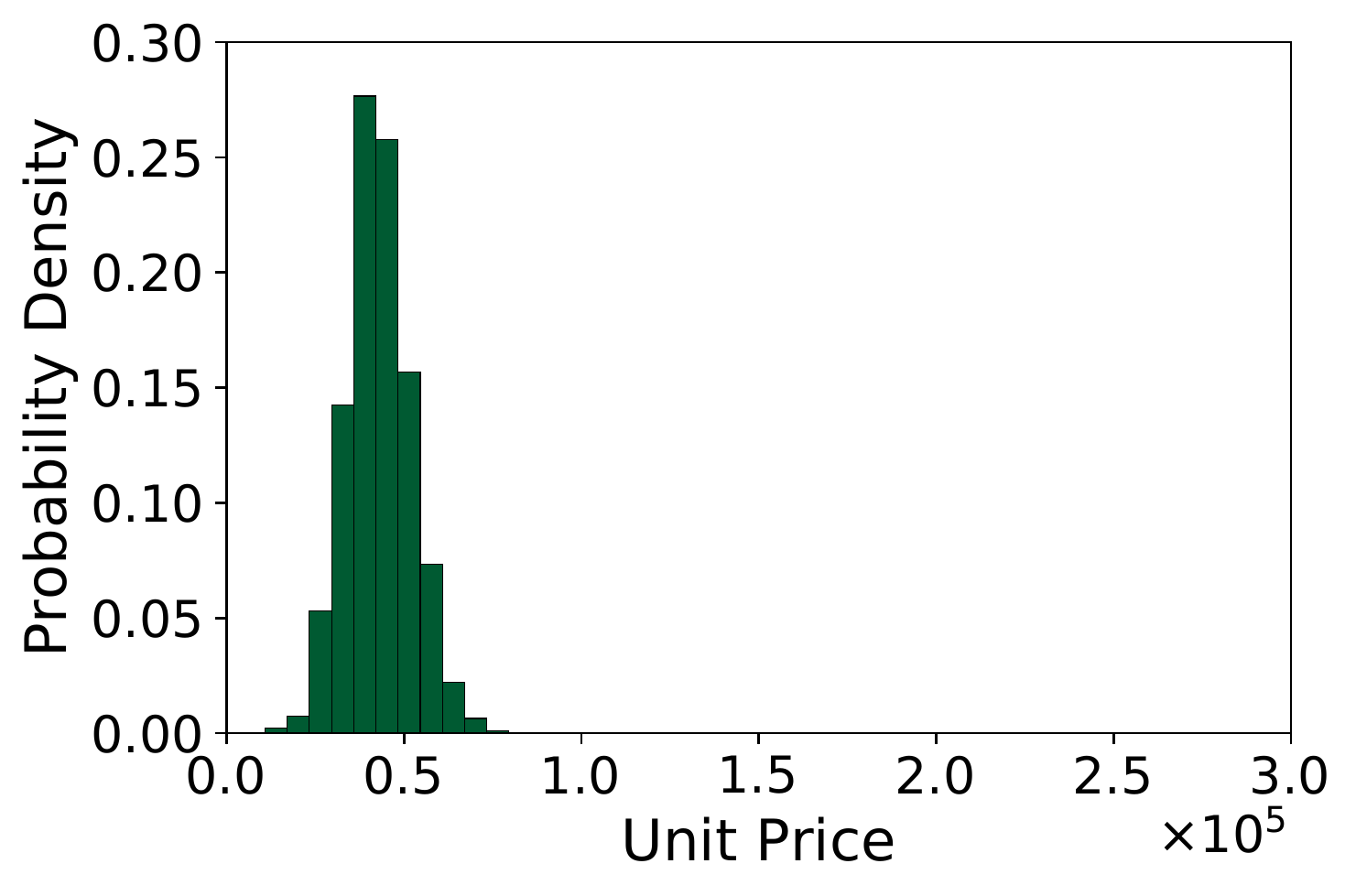}}
	\vspace{-2mm}
	\caption{Distributions of real estate unit prices in different urban districts.} 
	\vspace{-3mm}
	\label{fig:district_price}
\end{figure}
In general, a city is partitioned into several urban districts for administration. 
Each urban district can have distinctive urban functionalities and leads to diversified real estate price distributions, as illustrated in \figref{fig:district_price}. Inspire by \cite{zhu2016days}, we partition the tasks via urban districts, and each urban district corresponds to a learning task. These tasks share most of parameters of the model except have a unique fully-connected output layer to generate appraisal of distinctive distribution. Therefore, the real estate located in each urban district is valuated by
\begin{equation}
\hat{y}_{t+1}=FC_{m}(h_{t+1}^o),
\end{equation}
where $FC_m$ denotes the unique fully-connected layer of $m$-th task associated with $m$-th urban district.

Overall, our model aims to minimize the \textit{Mean Square Error}~(MSE) loss between the estimated unit price and the ground truth transaction unit price of real estate,
\begin{equation}
L = \frac{1}{|S|}\sum_{S_m \subset S}\sum_{s_{t+1} \in S_m}\left(\hat{y}_{t+1} - y_{t+1}\right)^2,
\end{equation}
where $S_m$ denotes the real estate transactions set happening in $m$-th urban district, $|S|$ denotes the cardinality of $S$.

\section{Experiments}\label{sec:exp}
\subsection{Experimental setup}
We evaluate the performance of \mugrep on both \beijing and \chengdu datasets. We chronologically order each dataset based on the transaction date. For \beijing, we take the data of which the transaction date ranges from January 2018 to June 2019, as the training set, the following one month as validation set, and the rest of data for testing. For \chengdu, the training set is set from January 2018 to June 2020, the other setting are the same as \beijing.
Our experiments mainly focus on (1)~the overall performance comparison, (2)~the ablation studies of model and features, (3)~the feature importance analysis, (4)~and the effectiveness check for spatially different communities.
Furthermore, please refer to Appendix \ref{sec:prototype} for the prototype system of real estate appraisal.

\subsubsection{\textbf{Implementation details.}}
We take $\epsilon_d=500$m, $\epsilon_{\tau}=90$, $N_e=5$ and $L_e=2$ for event-level representation learning. We choose $N_c=5$, set the distance thresholds~(\eg $\epsilon_g$) of several types of edge as the 0.001 quantile of all pair-wise euclidean distance values, and select $L_c=1$ for community-level representation learning. The dimensions of $h_{e,t+1}^{L_e}$, $h_u^i$ and $h_{c}^{i,L_c}$ are fixed to 32. The hidden dimension of $MLP$ is fixed to 64. We employ Adam optimizer, set learning rate as 0.01 to train our model, and early stop if the loss does not decrease on validation data set over 30 epochs. 

\subsubsection{\textbf{Evaluation metrics.}}
We adopt \textit{Mean Absolute Error}~(MAE), 
\textit{Mean Absolute Percentage Error}~(MAPE), and \textit{Root Mean Square Error}~(RMSE), three widely used metrics~\cite{wang2020crowdfunding} for evaluation.
Note that the estimated unit price and ground truth transaction unit price are in 10,000 CNY.

\subsubsection{\textbf{Baselines.}}
We compare \mugrep with one statistical baseline~(HA), three classic machine learning based baselines~(LR~\cite{pedregosa2011scikit}, SVR~\cite{pedregosa2011scikit} and GBRT~\cite{ke2017lightgbm}), and two artificial neural networks~(ANN) based baselines~(DNN, PDVM~\cite{bin2019peer}). The input features of all learning based methods are the same. The details of these baselines are introduced in Appendix \ref{sec:baselines}.


\begin{table}[tb]
	\vspace{-5mm}
	\centering
	\caption{Overall performance evaluated by MAE, MAPE, RMSE on \beijing and \chengdu datasets.}
	\vspace{-2mm}
	\setlength{\tabcolsep}{2.5mm}{
		\begin{tabular}{l|c|ccc}
			\toprule[0.8pt]
			\textbf{Dataset} & 
			\textbf{Algorithm} & \textbf{MAE} & \textbf{MAPE} & \textbf{RMSE}\\
			\midrule[0.5pt]
			\multirow{7}{*}{\beijing} & HA & 0.6313 & 11.33\%	& 1.1008 \\
			~ & LR & 0.4776	& 8.94\% & 0.7041\\
			~ & SVR & 0.4427 & 8.10\% & 0.6840\\
			~ & GBRT & 0.3640 & 6.70\% & 0.5515\\
			~ & DNN & 0.3550 & 6.35\% & 0.5505 \\
			~ & PDVM & 0.3469 &	6.17\% & 0.5373\\
			~ & \mugrep & \textbf{0.3244} & \textbf{5.76\%} & \textbf{0.5097}\\
			\midrule[0.5pt]
			\multirow{7}{*}{\chengdu} & HA & 0.1456	& 10.65\% & 0.2487 \\
			~ & LR & 0.1515	& 9.73\% & 0.2150 \\
			~ & SVR & 0.1339 & 8.57\% & 0.1888 \\
			~ & GBRT & 0.1133 & 7.27\% & 0.1708 \\
			~ & DNN & 0.1090 & 7.26\% & 0.1651 \\
			~ & PDVM & 0.1051 & 6.95\% & 0.1583 \\
			~ & \mugrep & \textbf{0.0916} & \textbf{6.20\%} & \textbf{0.1404} \\
			\bottomrule[0.5pt]
	\end{tabular}}
	\vspace{-3mm}
	\label{table:overall_results}
\end{table}	

\subsection{Overall Performance}
\tabref{table:overall_results} reports overall results of our methods and all compared baselines on two datasets with respect to three metrics. Overall, our model achieves the best performance among all the baselines. Moreover, we observe all ANN based algorithms~(DNN, PDVM, \mugrep) outperform the statistical algorithm~(HA) and machine learning based algorithms~(LR, SVR, GBRT), which consistently verifies the advantages of applying ANN to real estate appraisal for its extraordinary non-linear processing ability.
Particularly, \mugrep achieves $(6.5\%, 6.6\%, 5.1\%)$ and $(12.8\%, 10.8\%, 11.3\%)$ improvements beyond the state-of-the-art baseline PDVM for (MAE, MAPE, RMSE) on \beijing and \chengdu, respectively. The results demonstrate the effectiveness of \mugrep.
	
\subsection{Ablation Study}
In this section, we conduct ablation studies on \mugrep, including model ablation and feature ablation, to further verify the effectiveness of each component and feature group. The experiments are finished for three metrics on both \beijing and \chengdu datasets.

\subsubsection{\textbf{Model Ablation.}}
\begin{figure}[tb]
	\centering
	\vspace{-7mm}
	\hspace{-2mm}
	\subfigure[{MAE.}]{\label{fig:abmodel_mae}
		\includegraphics[width=0.33\columnwidth]{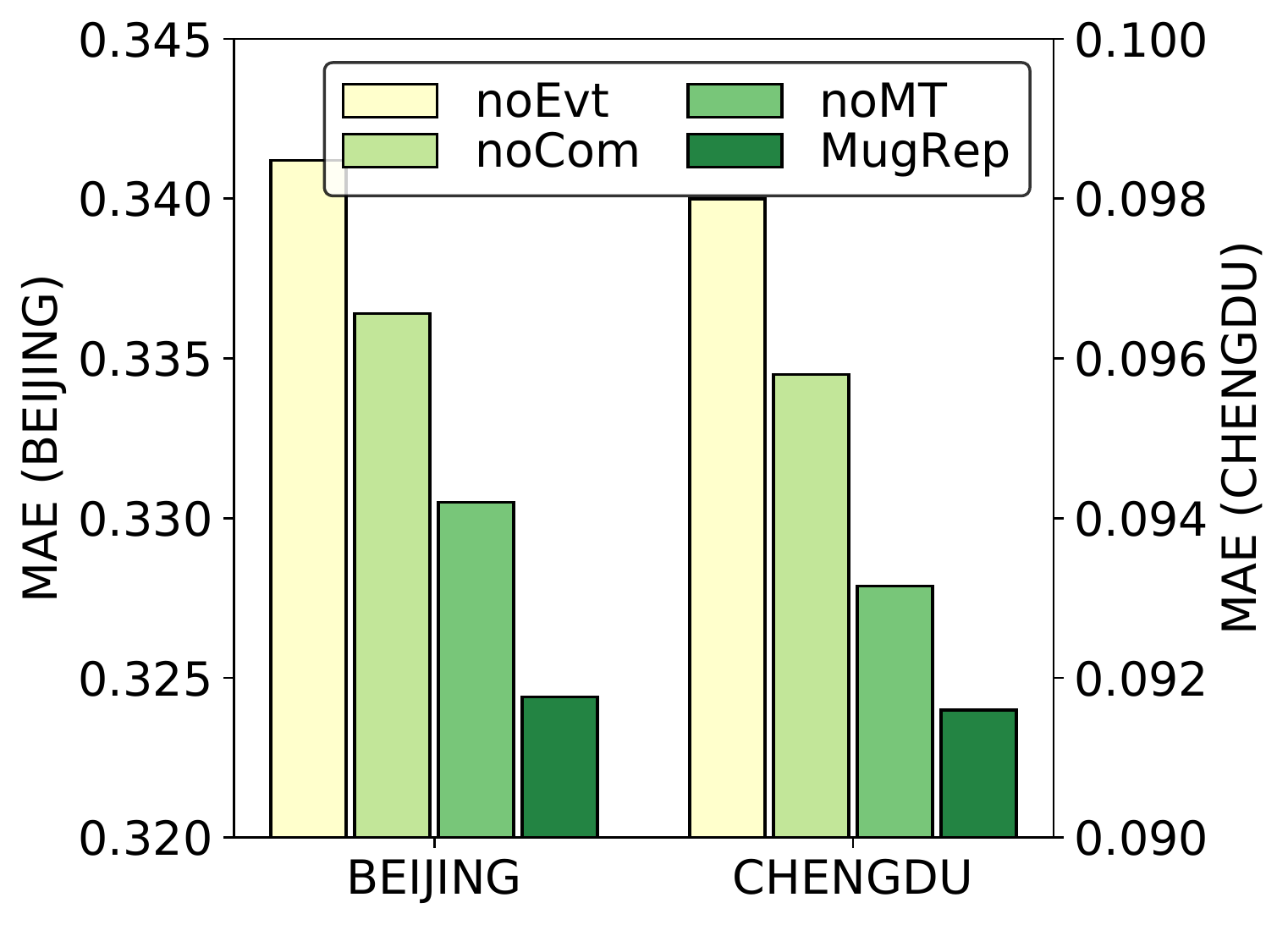}}\hspace{-1mm}
	\subfigure[{MAPE.}]{\label{fig:abmodel_mape}
		\includegraphics[width=0.33\columnwidth]{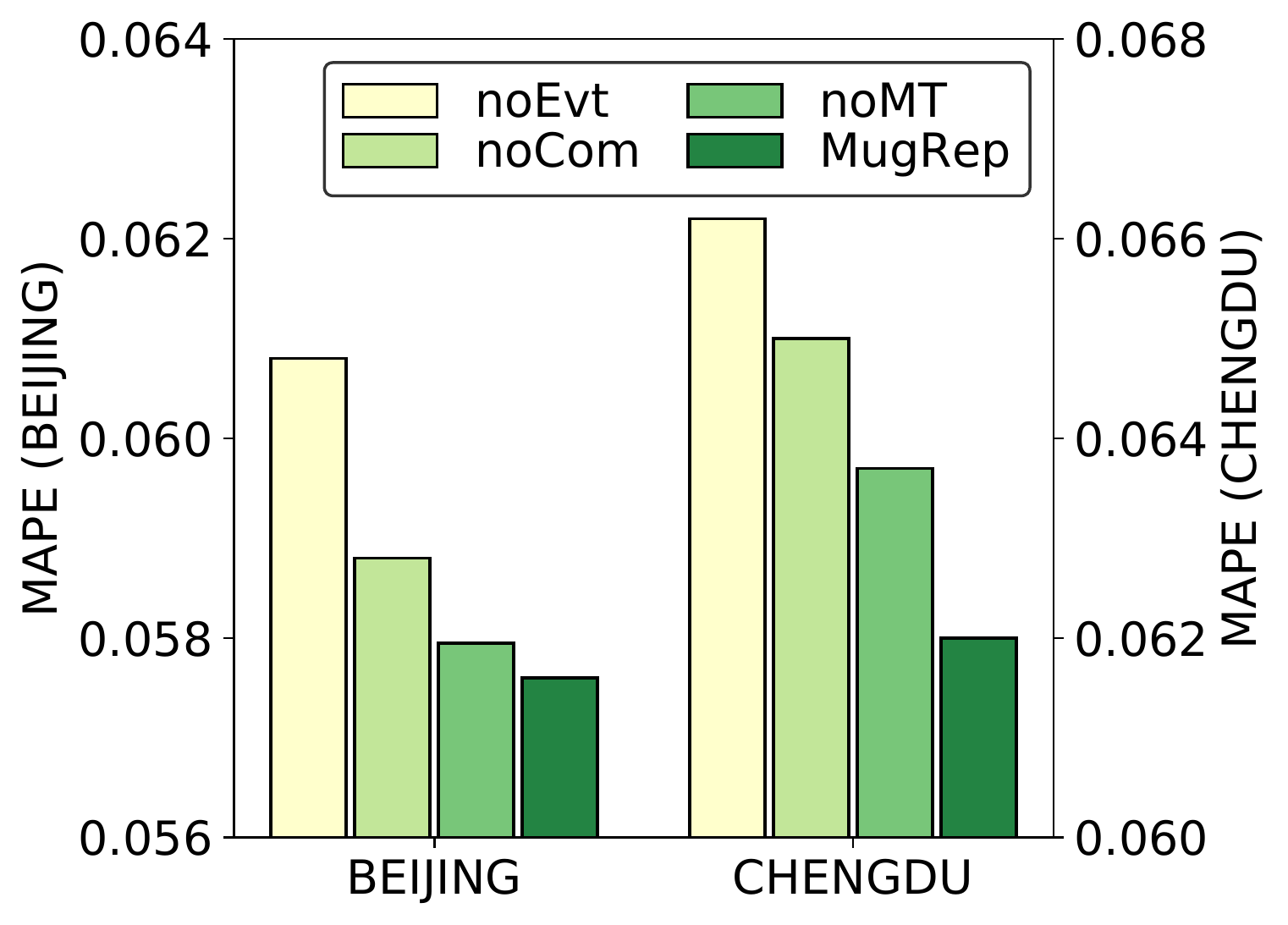}}\hspace{-1mm}
	\subfigure[{RMSE.}]{\label{fig:abmodel_rmse}
		\includegraphics[width=0.325\columnwidth]{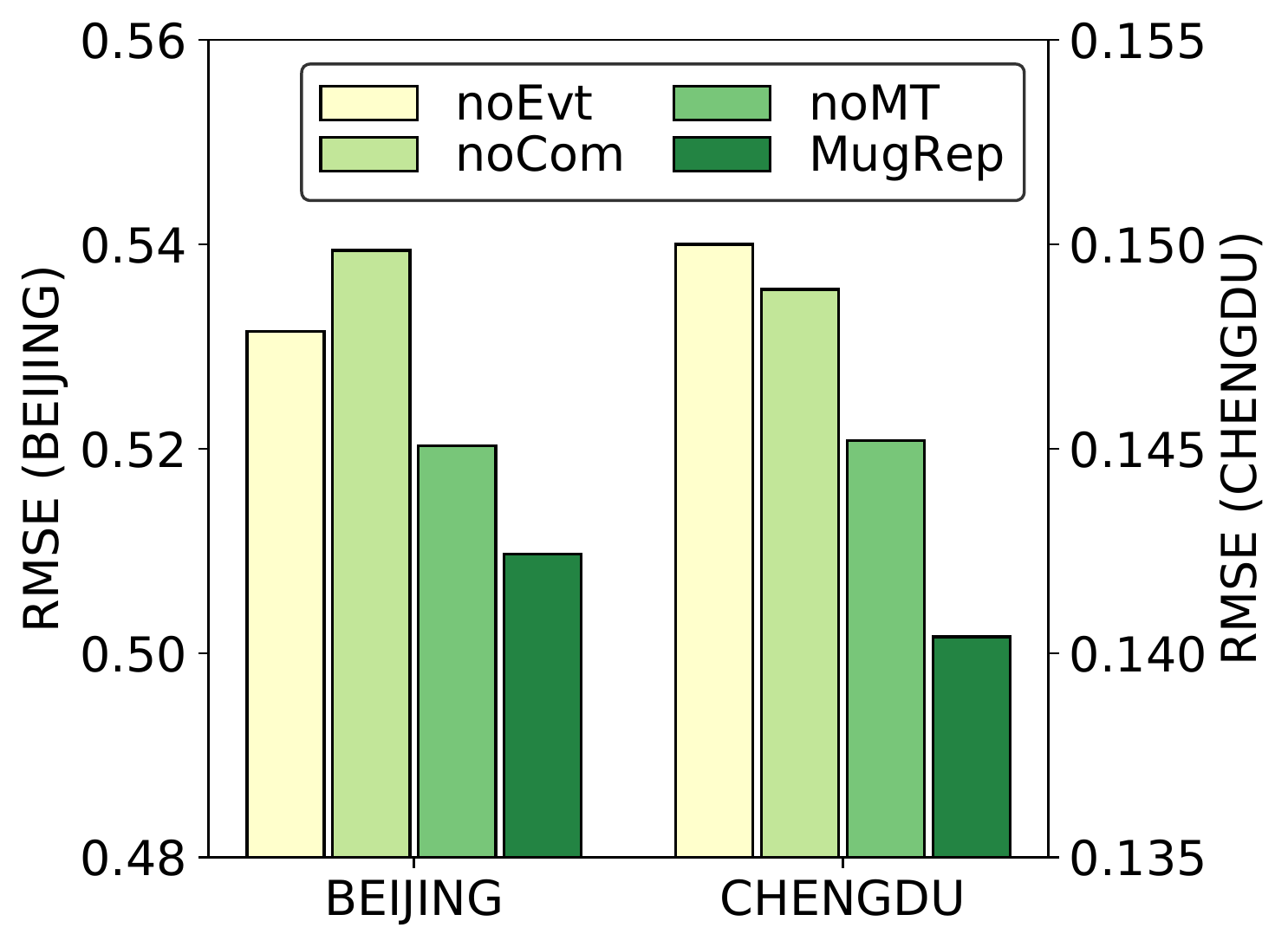}}
    \vspace{-2mm}
	\caption{Ablation tests of the model on two datasets.} 
	\vspace{-2mm}
	\label{fig:ablation_model}
\end{figure}

We evaluate the performance of \mugrep and it's three variants, which are (1)~\textbf{noEvt} removes the event-level representation learning module; (2)~\textbf{noCom} removes the community-level representation learning module; (3)~\textbf{noMT} removes the multi-task learning module. The ablation results are reported in \figref{fig:ablation_model}. As can be seen, removing any of the components leads to remarkable performance degradation. Among these components, we find the event-level and community-level representation modules are more important, especially the event-level representation module. This is because the nearby real estates have strong similarities and dependencies. The close historical real estate transactions can be a very valuable reference for subject property valuation.
All the results demonstrate the effectiveness of \mugrep and its each component.

\subsubsection{\textbf{Feature Ablation.}}
\begin{figure}[tb]
	\centering
	\hspace{-2mm}
	\subfigure[{MAE.}]{\label{fig:abfeat_map}
		\includegraphics[width=0.33\columnwidth]{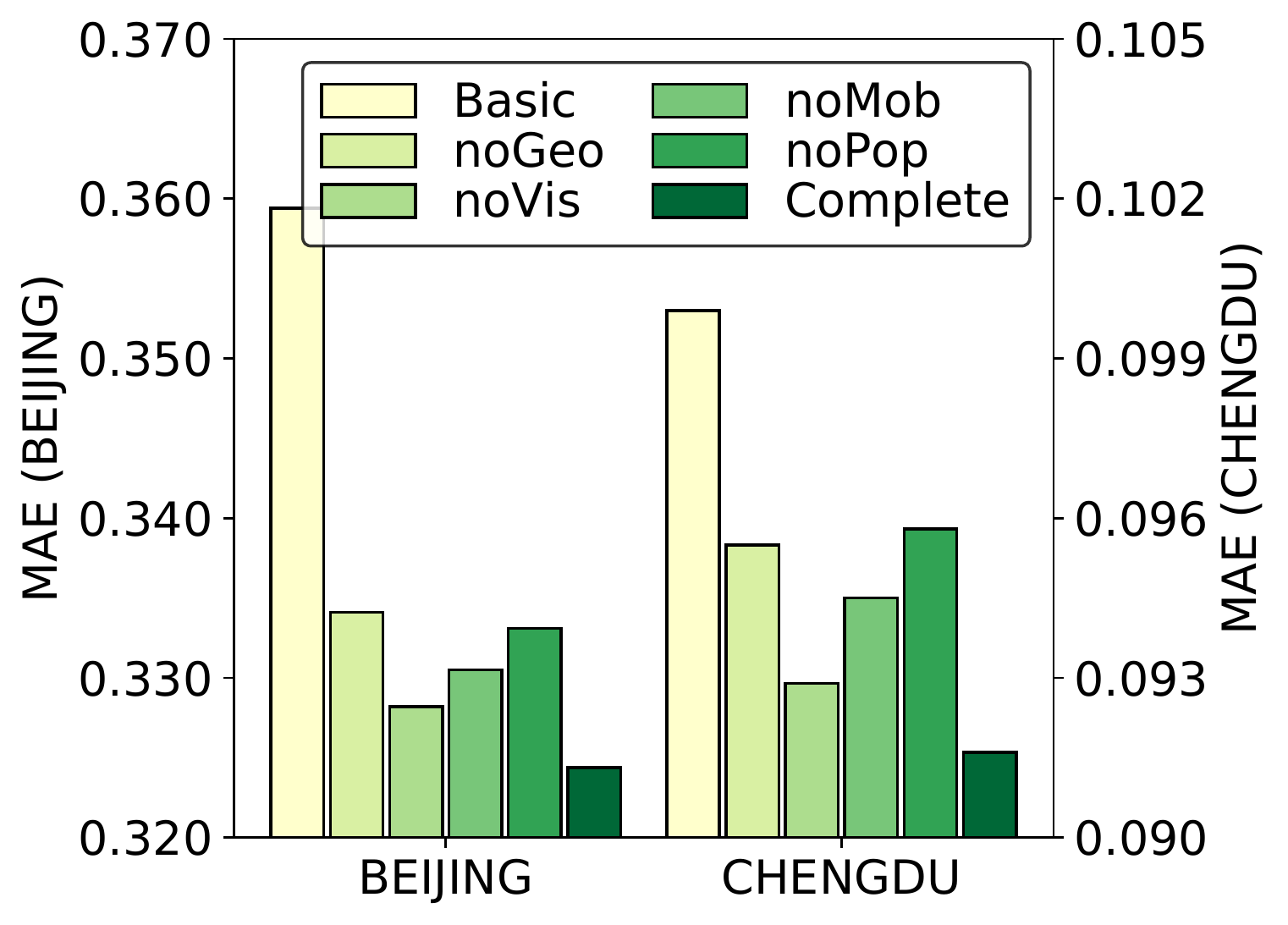}}\hspace{-1mm}
	\subfigure[{MAPE.}]{\label{fig:abfeat_mape}
		\includegraphics[width=0.33\columnwidth]{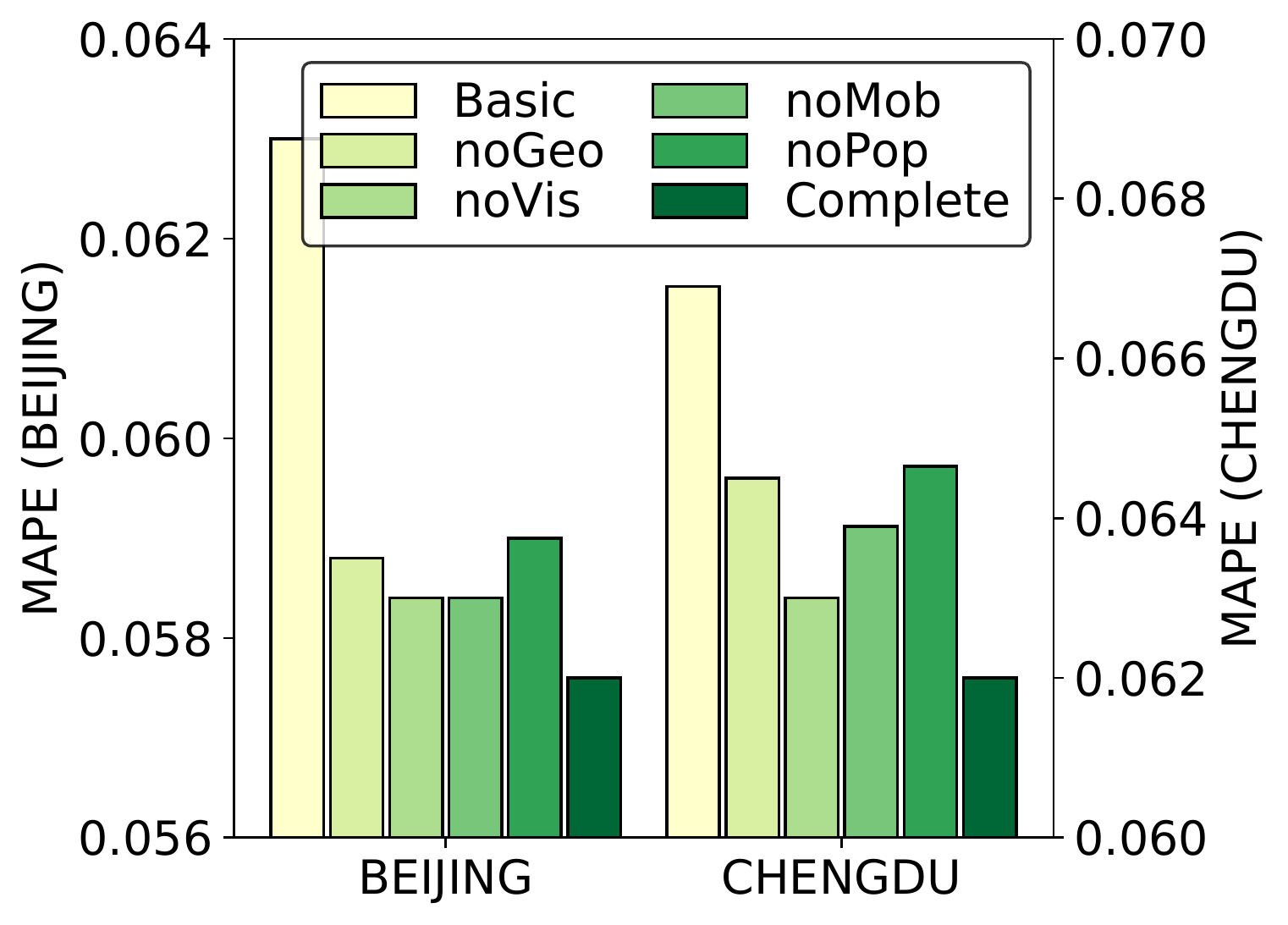}}\hspace{-1mm}
	\subfigure[{RMSE.}]{\label{fig:abfeat_rmse}
		\includegraphics[width=0.33\columnwidth]{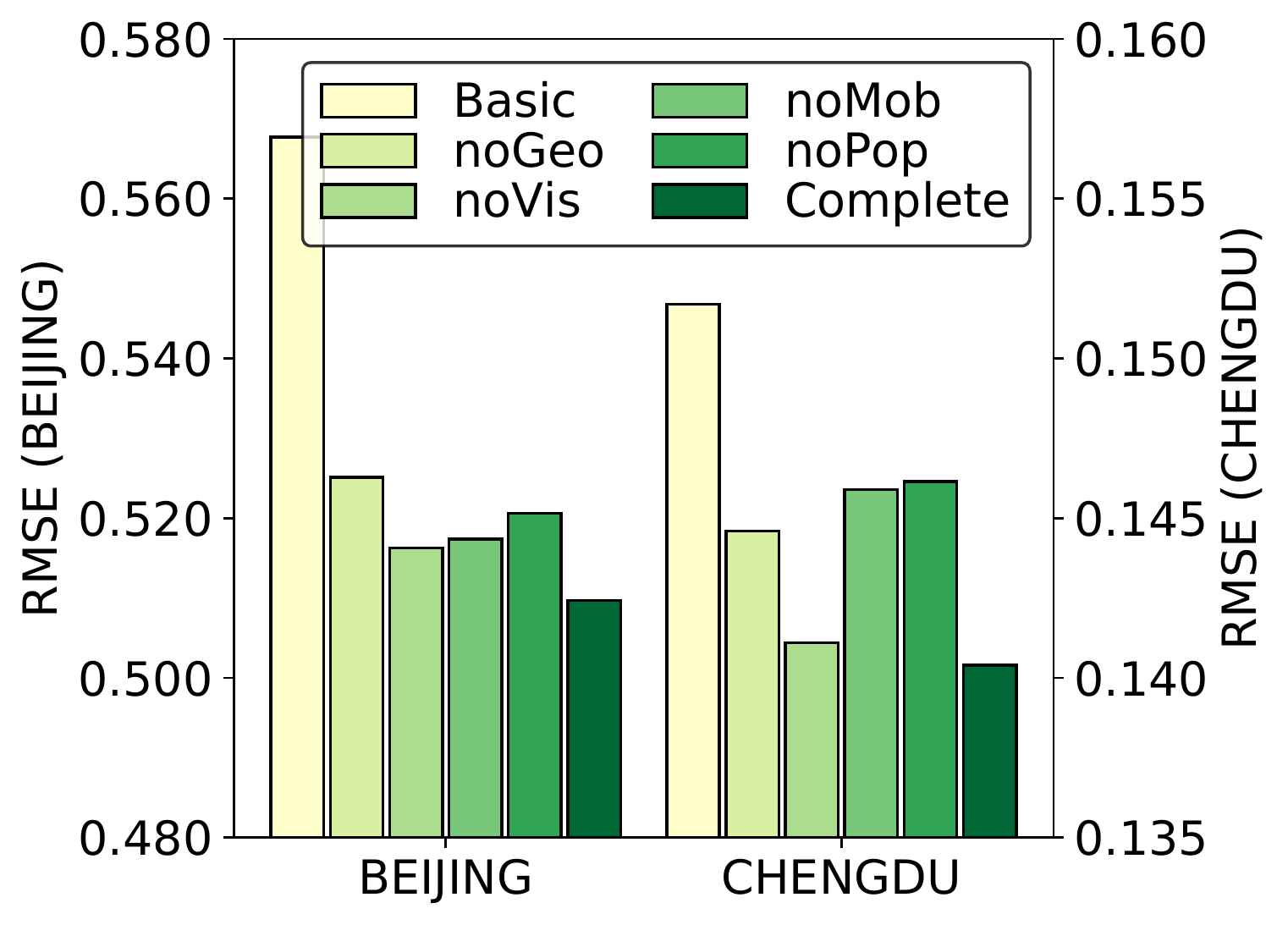}}
	\vspace{-2mm}
	\caption{Ablation tests of feature groups on two datasets.} 
	\vspace{-3mm}
	\label{fig:ablation_feature}
\end{figure}

To examine the performance impact of feature groups that constructed based on four additional multi-source urban datasets, we evaluate \mugrep with complete features and its five variants:~(1)~\textbf{Complete} is \mugrep with complete features; (2)~\textbf{Basic} excludes all features from four additional multi-source urban datasets and the community-level representation learning module; (3)~\textbf{noGeo} excludes geographical features and geographical edge set $\mathcal{E}_g$ in the heterogeneous inter-community graph; (4)~\textbf{noVis} excludes population visit features and corresponding edge set $\mathcal{E}_v$; (5)~\textbf{noMob} excludes mobility features and corresponding edge set $\mathcal{E}_m$; (6)~\textbf{noPop} excludes resident population profile features and corresponding edge set $\mathcal{E}_p$.
The experimental results are reported in \figref{fig:ablation_feature}. There is a consistent performance degradation by excluding any of additional urban feature groups. If we exclude all these additional urban features~(\ie basic), \mugrep will have significant $(10.8\%, 9.4\%, 11.4\%)$ and $(9.1\%, 7.9\%, 8.0\%)$ performance degradation for (MAE, MAPE, RMSE) on \beijing and \chengdu, which demonstrate the effectiveness of these urban features and community-level representation learning module. 
Besides, we observe \textbf{noMob} and \textbf{noPop} lead to notable performance degradation compared to \mugrep with complete features. The observation verifies that considering the characteristics and correlations of community residents are very useful for real estate appraisal.

\subsection{Feature Importance Analysis}
\begin{figure}[tb]
    \vspace{-5.5mm}
	\centering
	\includegraphics[width=0.98\columnwidth]{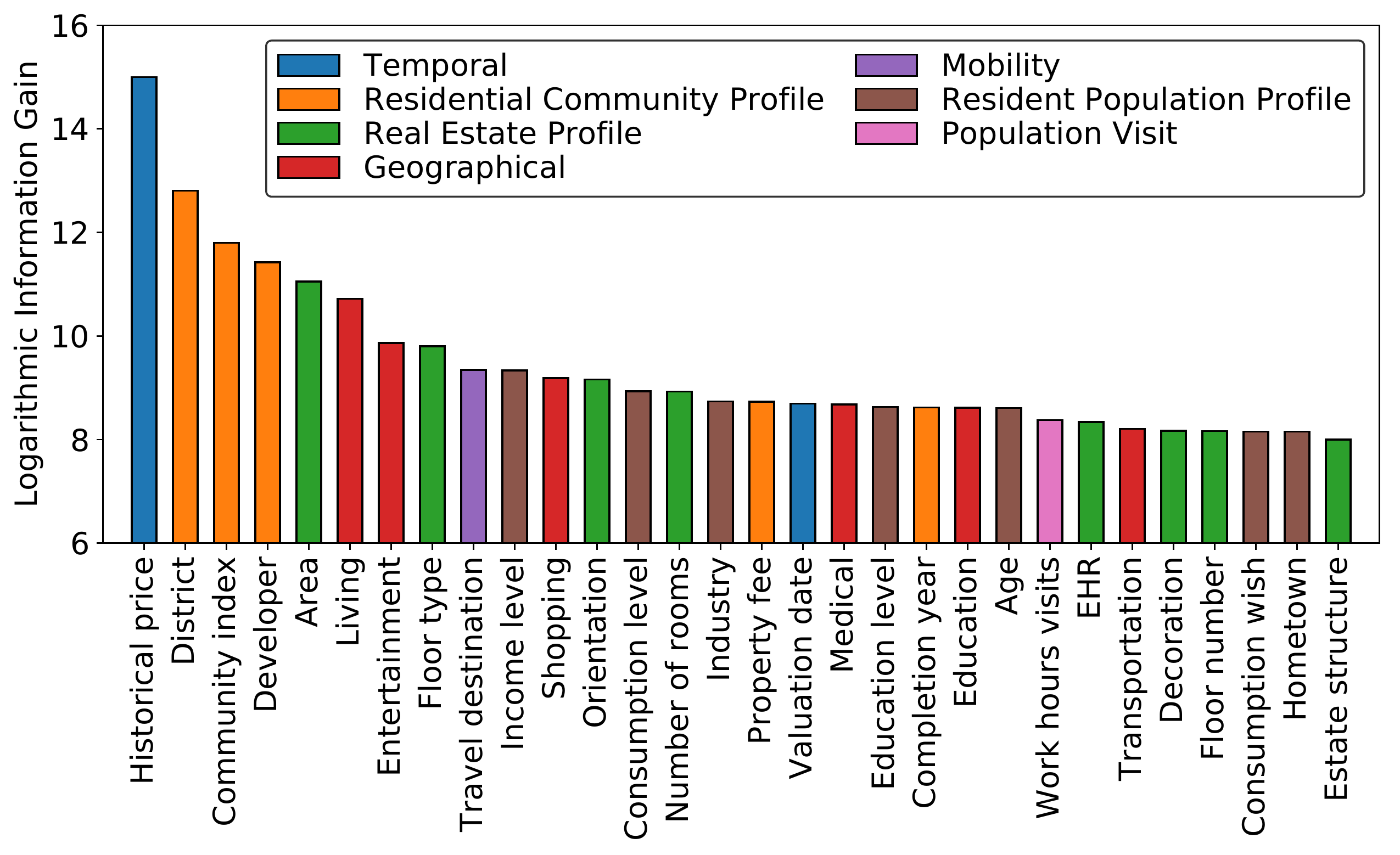}
	\vspace{-2mm}
	\caption{Top-30 features~(feature EHR refers to Elevator household ratio) ranked by logarithmic information gain.}
	\label{fig:feature_imp}
	\vspace{-3mm}
\end{figure}

To further evaluate the effect of our constructed features, we illustrate the top-30 most important features in \figref{fig:feature_imp}. The features are ranked by logarithmic information gain~\cite{ke2017lightgbm}. As can be seen, the distribution of historical price are the most important features. The subsequent one is the district of residential community, which indicates the large difference between districts. Furthermore, we observe half of the top-30 features are from the four multi-source urban datasets, which demonstrates the effectiveness of these urban features. 
Among these urban features, the geographical features~(Living, Entertainment, Shopping), mobility features~(Travel destination) and resident population profile features~(Income level, Consumption level, Industry) are ranked very high~(in top-15). For the reasons that these geographical features are closely related to living quality, these mobility features are greatly relevant to travel preferences of community residents, and these resident population profile features are strongly associated with the wealth of community residents. The living quality, travel preferences, and wealth are three very important factors to reflect the real estate prices.

\subsection{Effectiveness in Different Community}
\begin{figure}[tb]
	\centering
	\vspace{-6mm}
	\subfigure[{Distribution of MAPE.}]{\label{fig:area_mape}
		\includegraphics[width=0.48\columnwidth]{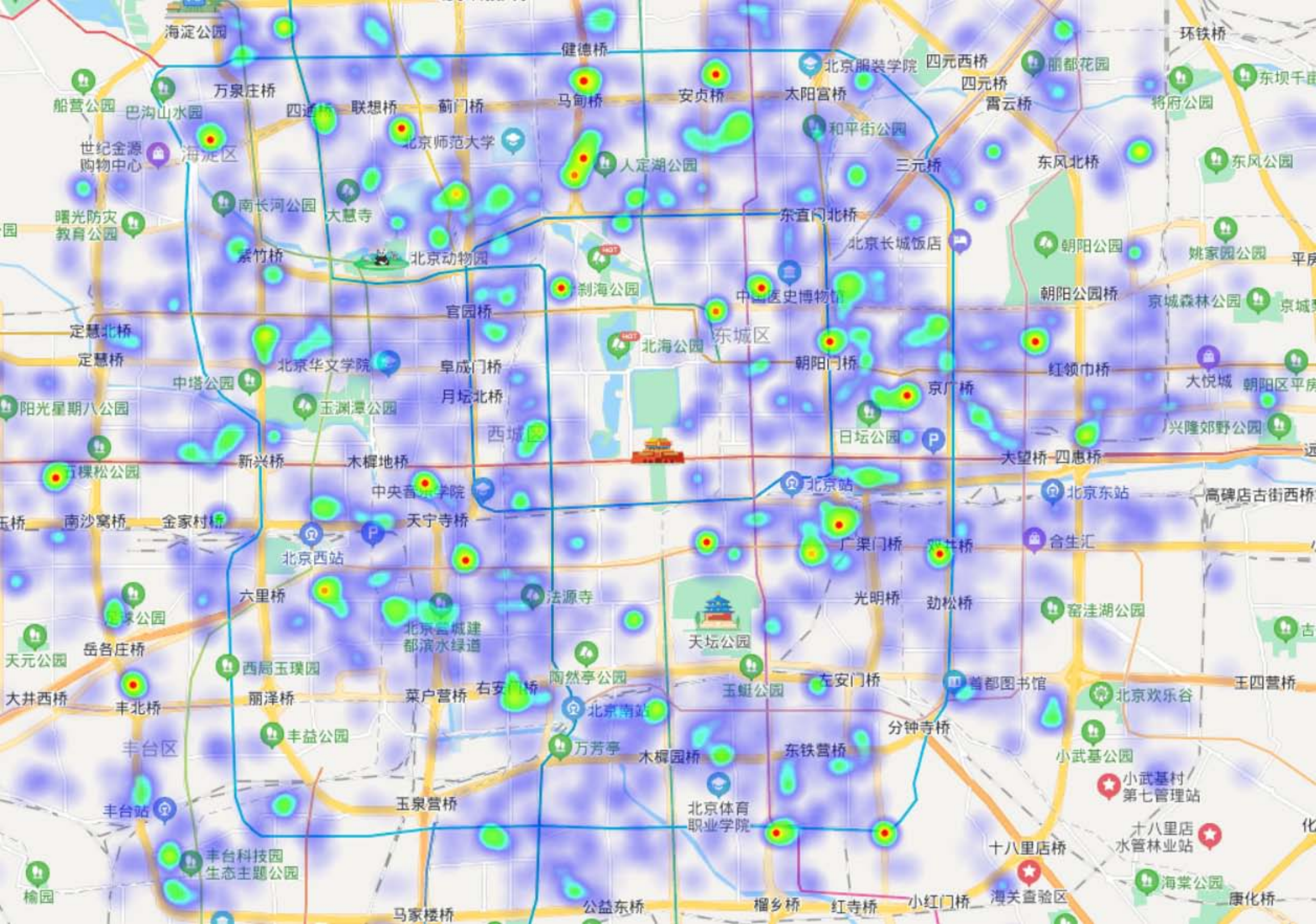}}
	\subfigure[{Distribution of inverse transaction volume.}]{\label{fig:area_volume}
		\includegraphics[width=0.48\columnwidth]{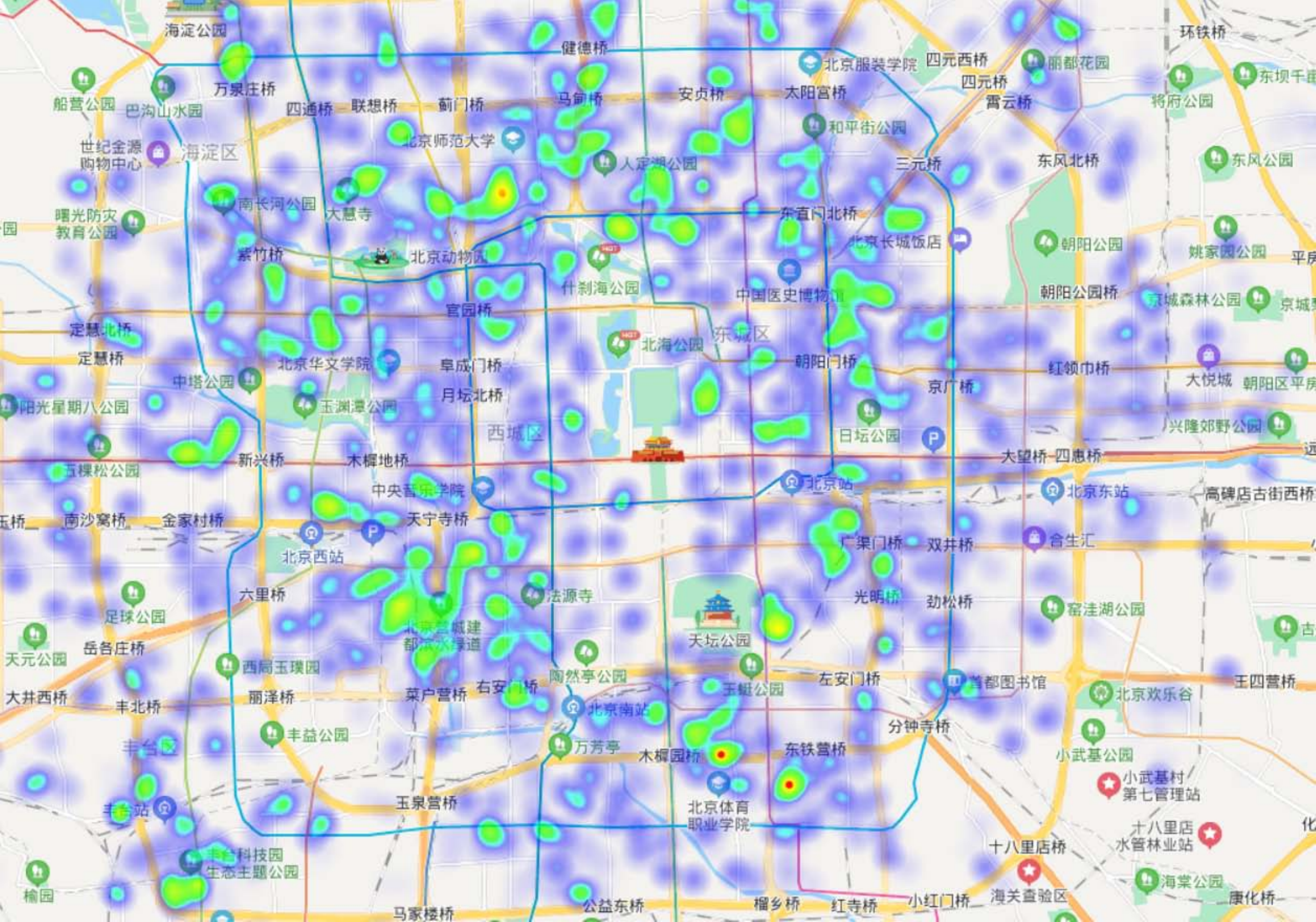}}
	\vspace{-2mm}
	\caption{Spatial distributions of MAPE and inverse transaction volume on \beijing} 
	\vspace{-3mm}
	\label{fig:area_effect}
\end{figure}

To evaluate the performance of \mugrep on spatially diverse communities, we compute the separate MAPE for each residential community. \figref{fig:area_mape} and \figref{fig:area_volume} show the spatial distribution of MAPE and inverse transaction volume on \beijing. As can be seen, some communities with high MAPE~(bright color) always have high inverse transaction volume~(\ie less transaction volume) nearby. This makes sense for two reasons: first, the historical price features of same community are important for subject property valuation, less transaction volume in the community means inaccurate or missing historical price features; second, the evolving transaction event graph convolution module of \mugrep is highly correlated with nearby previous transactions, the absence of nearby transactions decreases the effectiveness of event-level representation. This result indicates further optimization can be applied to these residential communities with scarce transactions to improve the overall performance.

\section{Related work}\label{sec:related}
\noindent \textbf{Real Estate Appraisal.}
Traditional real estate appraisal methodologies can be mainly categorized into three classes, \ie sales comparison based approaches~\cite{Mccluskey1997AnEO}, cost based approaches~\cite{guo2014integrated}, and income based approaches~\cite{baum2017income}.
The sales comparison approach estimates the market value of real estate depending on some forms of comparison.
The cost approach is based on the theory that the summation of land value and depreciated value of any improvements can be as the estimation for the real estate. 
And the income approach estimates the real estate market value based on its income.
Hedonic price model~\cite{rosen1974hedonic,cheshire1995price} is also commonly used in real estate appraisal. It assumes that the real estate can be viewed as an aggregation of individual attributes, which implicitly reflect the real estate market value. However, it does not consider attributes interactions and is incompetent in non-linear data \cite{limsombunchai2004house}.
Besides, automated valuation methods~(AVMs) have arisen researchers' interests for they can automatically estimate the market value of an real estate based on its available attributes. Many AVMs such as linear regression \cite{csipocs2008linear,ahn2012using}, support vector regression \cite{lin2011predicting}, boosted regression trees \cite{graczyk2010comparison, park2015using} and artificial neural networks methods~\cite{peterson2009neural,peter2020review,poursaeed2018vision, you2017image,law2019take} are widely applied into the real estate appraisal.
Moreover, some works \cite{fu2014exploiting, fu2014sparse, fu2015real} investigate how to rank real estates via various viewpoints, such as individual, peer and zone dependency, online user reviews and offline moving behaviors, and diverse mixed land use. 
Furthermore, there are few works that try to capture the peer-dependency among nearby estates.
\citet{fu2014exploiting} use the generative likelihood of each edge to model peer-dependency, which does not adequately integrate the attributes knowledge of nearby estates.
Works~\cite{bin2019peer,you2017image} sample fixed number of similar estates by selecting k-nearest similar estates or the random walk algorithm, and then feed these samples to recurrent neural networks. The sampling process could lead to the loss of valuable information.
Overall, these prior studies all leave out community residents characteristics, and are incapable to fully model the spatiotemporal dependencies among real estate transactions. 
Besides, none of them attempt to capture the diversified correlations between residential communities.
	
\noindent \textbf{Real Estate Forecasting}. This work is also related to real estate forecasting tasks. 
For example, \citet{tan2017time} proposes a time-aware latent hierarchical model and \citet{ge2019integrated} proposes an integrated framework that improving the DenseNet to predict future house prices of regions.
\citet{zhu2016days} proposes a multi-task
linear regression model for real estate's days-on-market prediction.
\citet{li2019housing} estimates the housing demand based on express delivery data.

\noindent \textbf{Graph Neural Network.}
Graph neural networks~(GNN) is designed to extend the well-known convolution neural network to non-Euclidean graph structures. GNN is usually used to obtain the expressive representation of each node by aggregating and transforming the representations of its neighbors in the graph~\cite{kipf2017semi,hamilton2017inductive,velivckovic2017graph}. 
Many previous studies have demonstrated the effectiveness of GNN in the graph-based representation learning tasks~\cite{wu2020comprehensive}.
Because of its effectiveness, GNN has been successfully applied to various fields, such as physics~\cite{santoro2017simple}, chemistry~\cite{gilmer2017neural}, biology~\cite{fout2017protein}, recommendation systems~\cite{ying2018graph,wang2019knowledge,xin2021out}, and smart city systems~\cite{zhang2020semi,liu2020multi,han2020joint}.
However, to the best of our knowledge, this is the first work applying GNN to real estate appraisal.


\section{Conclusion}\label{sec:conclusion}
In this paper, we presented \mugrep, a multi-task hierarchical graph representation learning framework for real estate appraisal. We first constructed abundant valuable features with respect to the fundamental attributes and community residents characteristics to comprehensively profile the real estate value. Then we designed an evolving transaction event graph convolution module to model the asynchronously spatiotemporal dependencies among real estate transactions, and devised a hierarchical heterogeneous community graph convolution module to capture diversified correlations between residential communities.
After that, an urban district partitioned multi-task learning module was introduced to perform the real estate appraisal of distinctive distribution. 
Extensive experiments on two real-world datasets demonstrated that \mugrep achieves the best performance compared with six baselines.

\begin{acks}
    This research is supported in part by grants from the National Natural Science Foundation of China (Grant No.91746301, 71531001).
\end{acks}

\bibliographystyle{ACM-Reference-Format}
\balance
\bibliography{sample-base}

\appendix
\balance
\begin{table*}[tb]
	\begin{small}
	\centering
	\vspace{-5mm}
	\caption{Detailed feature list.}
	\vspace{-3mm}
	\setlength{\tabcolsep}{1mm}{
		\begin{tabular}{l|l|l}
			\toprule[0.8pt]
			\textbf{Feature Type} & \textbf{Feature} & \textbf{Description}\\
			\midrule[0.5pt]
			\multirow{12}{*}{\tabincell{l}{Real Estate\\ Profile}} & Number of rooms & Number of rooms~(\eg bedroom, living-room, kitchen, bathroom) in this estate.\\
			~ & Area & Number of square meters of the estate.\\
			~ & Decoration & Type of decoration (\eg simply-decorated, well-decorated, not decorated).\\
			~ & Orientation & Orientation of the estate (\eg north, south, west, east).\\
			~ & Estate structure & Structure of estate (\eg flat layer, jump layer, duplex).\\
			~ & Heating method & Method of estate's heating~(\eg~central heating, self-heating, without heating).\\ 
			~ & Floor type & Type of the floor of the estate (\eg high, medium, low, basement).\\
			~ & Free of tax & Whether the estate is free of sales tax.\\
			~ & Transaction ownership & Transaction ownership of the estate~(\ie~commercial housing, affordable housing, purchased public housing). \\
			~ & Floor number & Number of floors of the building where the estate is located.\\
			~ & Building type & Type of the building~(\eg tower building, slab-type building, mixed-type building).\\
			~ & Elevator household ratio & Ratio of the number of elevators to households of the building.\\ 
			\midrule[0.5pt]
			\multirow{7}*{\tabincell{l}{Residential\\ Community\\ Profile}}  & Developer & Developer of the community.\\ 
			~ & Completion year & Completion year of the community.\\ 
			~ & Number of buildings & The total number of buildings in the community.\\ 
			~ & Number of estates & The total number of estates in the community.\\ 
			~ & Property fee & Property fee of the community.\\ 
			~ & District & District of the community~(\eg~Xicheng, Haidian, Chaoyang).\\ 
			~ & Community index & Index of community.\\
			\midrule[0.5pt]
			\multirow{3}*{Temporal} & Valuation date & Date of making valuation for the estate.  \\
			~ & Historical price & \tabincell{l}{Statistics~(\eg mean, variance, maximum, minimum) for the unit prices of transactions closed in previous\\ quarter in the community.}\\
			\midrule[0.5pt]
			\multirow{10}{*}{Geographical} & Transportation & Number of the transport facilities~(\eg subway and bus stations) nearby and the distances of the nearest ones.\\
			~ & Education & \tabincell{l}{Number of the educational facilities~(\eg kindergarten, primary and middle schools, college) nearby\\ and the distances of the nearest ones.}\\
			~ & Medical & Number of the medical facilities~(\eg hospital, clinic, pharmacy) nearby and the distances of the nearest ones.\\
			~ & Shopping & \tabincell{l}{Number of the shopping places~(\eg mall, supermarkte, convenience store) nearby and the distances\\ of the nearest ones.}\\
			~ & Living & Number of the living places~(\eg restaurant, barbershop, bank) nearby and the distances of the nearest ones.\\
			~ & Entertainment & Number of the entertainment venues~(\eg cinema, gym, park) nearby and the distances of the nearest ones.\\
			~ & Unpleasantness & Number of the unpleasant facilities~(\eg factory, cemetery) nearby and the distances of the nearest ones.\\
			~ & Number of facilities & Number of all the POIs and transport stations nearby.\\
			\midrule[0.5pt]
			\multirow{3}*{Population Visit} & Work hours visits & Visiting frequency of population nearby in work hours~(10:00-18:00) on workdays and weekends. \\
			~ & Break hours visits & Visiting frequency of population nearby in break hours~(18:00-23:00) on workdays and weekends. \\
			~ & All day visits & Visiting frequency of population nearby in all day on workdays and weekends.\\
			\midrule[0.5pt]
			\multirow{6}{*}{Mobility} & Inflow volume & Human volume of the inflow of community on workdays and weekends.\\ 
			~ & Outflow volume & Human volume of the outflow of community on workdays and weekends.\\ 
			~ & Travel mode & 
			\tabincell{l}{Distribution of the residents' travel modes~(\eg drive, taxi, bus, cycle, walk) from the community on\\ workdays and weekends.}\\  
			~ & Travel destination & \tabincell{l}{Distribution of the types of residents' travel destinations~(\eg enterprise, administration, shopping places,\\ entertainment venues) from the community on workdays and weekends.}\\ 
			\midrule[0.5pt]
			\multirow{12}*{\tabincell{l}{Resident\\ Population\\ Profile}} & Resident population & Number of resident population in the community. \\
			~ & Hometown & Distribution of the hometowns~(\eg Beijing, Shandong) of resident population in the community.\\
			~ & Gender & Distribution of the gender of resident population in the community.\\
			~ & Age & Distribution of the ages~(\eg teenager, youth, the middle-aged, the old) of resident population in the community.\\
			~ & Life stage & Distribution of the life stages~(\eg student, working, parent, retire) of resident population in the community.\\
			~ & Industry & Distribution of the industries~(\eg education, catering, IT, finance) of resident population in the community.\\
			~ & Car owner & Distribution of owning cars of resident population in the community.\\
			~ & Income level& Distribution of the income levels~(\eg low, medium, high, very high) of resident population in the community.\\
			~ & Education level & Distribution of the education levels~(\eg undergraduate, college, senior) of resident population in the community.\\
			~ & Consumption level & Distribution of the consumption level~(\eg low, medium, high) of resident population in the community.\\
			~ & Consumption wish & \tabincell{l}{Distribution of the consumption wishes~(\eg daily supplies, education, healthcare, travel, finance, technology)\\ of resident population in the community.}\\
			\bottomrule[0.5pt]
	\end{tabular}}
	\label{table:features}
	\end{small}
\end{table*}
\newpage

\section{Details of used features} \label{sec:feature_detial}
 \tabref{table:features} is the feature list used in \mugrep, including Real Estate Profile features, Residential Community Profile features, Temporal features, Geographical features, Population Visit features, Mobility features, and Resident Population Profile features.

\section{Details of baselines} \label{sec:baselines}
We compare our \mugrep with the following six baselines.
We carefully tune major hyper-parameters of each baseline based on their recommended settings. GBRT, DNN, and PDVM employ the same early stop training strategy as \mugrep.
\begin{itemize}
	\item \textbf{HA} uses the average previous 90 days' price of transactions closed in the same residential community as estimated value.  
	\item \textbf{LR}~\cite{pedregosa2011scikit} makes appraisal via the well-known linear regression model. 
	\item \textbf{SVR}~\cite{pedregosa2011scikit} makes appraisal via the support vector regression model. We use the Radial Basis Function~(RBF) kernel.
	\item \textbf{GBRT} makes appraisal via gradient boosted regression tree model. We use the version in LightGBM~\cite{ke2017lightgbm}, set learning rate to 0.1, set maximal tree depth to 10, and maximal leaves number to 31. 
	\item \textbf{DNN} is a type of ANN method contains two fully-connected 64 dimensions hidden layers with ReLU activation functions, and employ Adam for optimization. The learning rate is set to 0.005.
	\item \textbf{PDVM}~\cite{bin2019peer} is a state-of-the-art ANN method for real estate appraisal. It models estates peer-dependency by using the k-nearest similar estate sampling to sample fixed number of real estates, and feed them to a bidirectional LSTM to generate final real estate appraisal. We slightly modify PDVM to sample historical real estate transactions to fit our dataset.
	We employ one layer bidirectional LSTM, and its input sequence length is 7. The hidden dimension is set to 64, learning rate is 0.005.
\end{itemize}

\begin{figure}[htb]
	\centering
 	\vspace{2mm}
	\includegraphics[width=0.99\columnwidth]{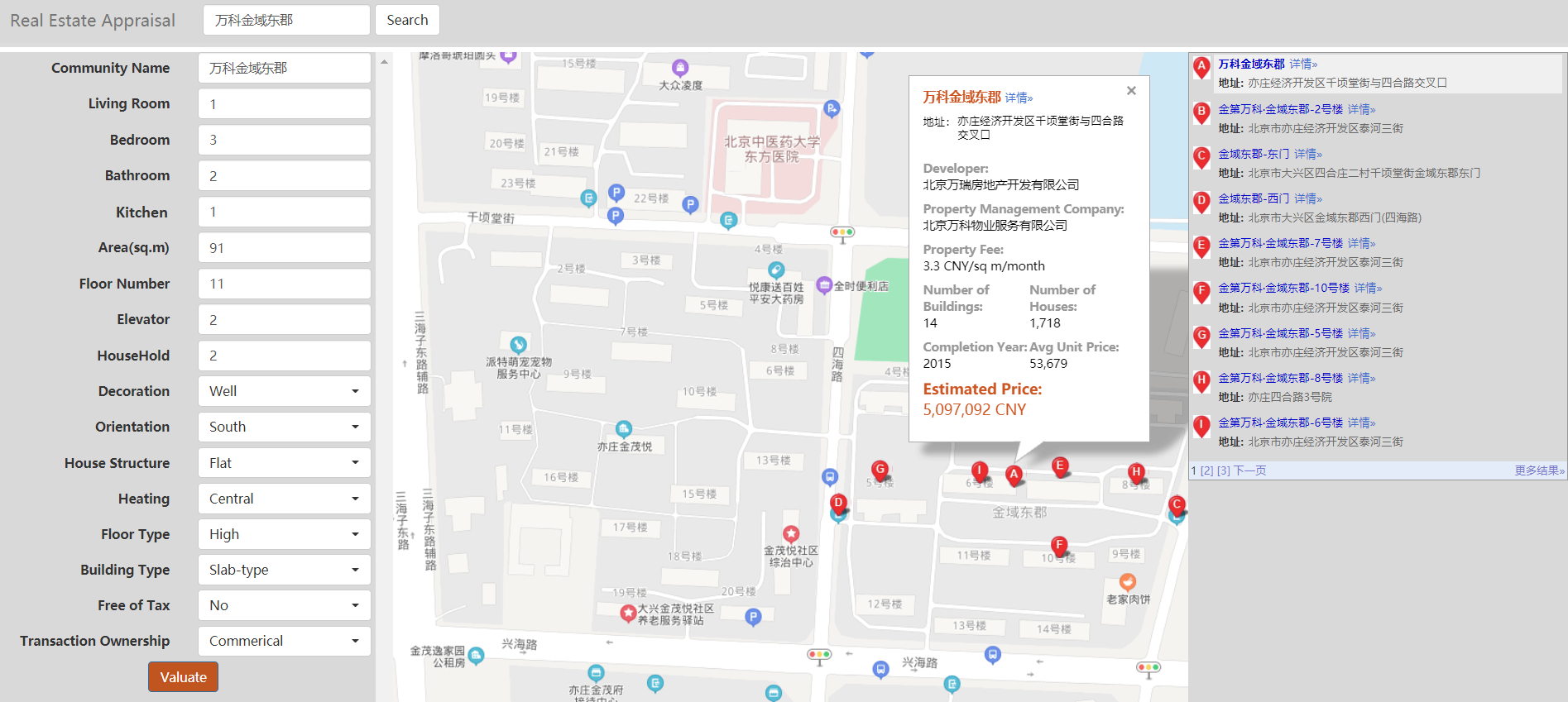}
	\caption{Prototype system.}
	\label{fig:prototype}
	\vspace{-2mm}
\end{figure}

\section{Prototype system}\label{sec:prototype}
We have implemented a prototype system for supporting users to make real estate appraisal decisions. 
We use angularJS~(JavaScript MVW framework), bootstrap~(front-end framework for web development), and Django~(a web framework in Python) along with MySQL to build our prototype system, of which \figref{fig:prototype} shows a screenshot.
Specifically, once the user enters a residential community name to search, and select the community from a list of returned candidate items, the system will show its position on the map and other detailed information, including developer, property fee, completion year, \etc
Then the user is expected to input real estate profile attributes, such as estate's number of rooms, area, decoration, orientation, and click the "Valuate" button to generate estimated price of the subject property.

\end{document}